\documentclass[10pt,twocolumn,letterpaper]{article}

\usepackage[pagenumbers]{cvpr} 

\usepackage{amsmath}
\usepackage{amssymb}
\usepackage{mathtools}
\usepackage{amsthm}
\usepackage[table,xcdraw]{xcolor}
\usepackage{pifont}
\usepackage{multirow}
\usepackage{textcomp}

\usepackage{graphicx}

\definecolor{cvprblue}{rgb}{0.21,0.49,0.74}
\usepackage[pagebackref,breaklinks,colorlinks,allcolors=cvprblue]{hyperref}

\def\confName{CVPR}
\def\confYear{2027}

\title{Imagine Smarter, Drive Faster:\\Efficient World Modeling for Decision-Aware Autonomous Driving}

\author{
Feiyang Jia$^{1,\dagger}$,
Lin Liu$^{2,\dagger}$,
Zhicheng Bao$^{1,\dagger}$,
Caiyan Jia$^{1,\ast}$,\\
Ziying Song$^{3,\ast}$,
Shengkai Zhang$^{1}$,
Lina Han$^{1}$,
Yan Zhao$^{1}$,
Xiaoshuai Hao$^{4,\ddagger}$\vspace{1ex}\\
$^{1}$Beijing Jiaotong University\quad
$^{2}$Dalian University of Technology\\
$^{3}$Nanyang Technological University\quad
$^{4}$Li Auto Inc.\\
}

\makeatletter
\apptocmd{\@maketitle}{%
  \begin{minipage}{\textwidth}
    \centering
    \includegraphics[width=\linewidth]{fig/fastworld-motivation_v3.pdf}
    \captionof{figure}{
\textbf{Rethinking world modeling for efficient autonomous driving.} 
(a) Representative planners exhibit substantial gaps between their best candidate and selected trajectories, suggesting that candidate selection, beyond representation capacity, is a key planning bottleneck.
(b) FastWorld pursues efficiency through a natively lightweight architecture combining compact representations, short-horizon future prediction, and selective trajectory refinement, without teacher distillation or additional domain-specific pre-training.
(c) Compared with existing E2E-based methods~\cite{kirby2026driving_DrivoR,xing2025goalflow,li2024hydra,liao2025diffusiondrive,wang2026meanfuser,chitta2022transfuser} and World-Model-based methods~\cite{li2025end_wote,wang2026drivejepa}, FastWorld achieves a favorable performance--efficiency balance on NAVSIMv1~\cite{dauner2024navsim}, delivering 92.2 PDMS at 132 FPS with a single 4090 GPU and only 27M parameters.
    }
    \label{fig:motivation}
  \end{minipage}
  \par\vspace{1em}
}{}{}
\makeatother

\begin{document}
\maketitle 
\begingroup
\renewcommand{\thefootnote}{}
\footnotetext{$\dagger$ Equal contribution. $\ast$ Corresponding author. $\ddagger$ Project lead.}
\endgroup

\begin{abstract}
World models hold great promise for autonomous driving planning, yet their ability to imagine the future often comes at a formidable computational cost. Efficiency efforts, such as knowledge distillation and domain-specific pre-training, often shift this burden to costly teachers or training pipelines, leaving lightweight world modeling an open challenge. We introduce \textbf{FastWorld}, a natively lightweight world-model framework for efficient autonomous driving planning. FastWorld reduces computation through compact scene representations, short-horizon action-conditioned prediction, and selective candidate verification and score refinement. These designs support efficient future-informed planning without teacher distillation or additional domain-specific pre-training. On NAVSIMv1, FastWorld achieves \textbf{92.2 PDMS at 132 FPS} on a single 4090 GPU. It also achieves 89.4 EPDMS on NAVSIMv2 and an average L2 of 0.40 over 3 seconds at 17 FPS on nuScenes. Code and models can be found at \url{https://github.com/MoyangSensei/FastWorld}.
\end{abstract}

\section{Introduction}
\label{sec:intro}

Autonomous driving has evolved from directly mapping sensor inputs to driving actions toward anticipatory decision-making that predicts environmental dynamics and evaluates action consequences. World models provide an important foundation for this transition~\cite{tu2025role_dwm_survey,guan2024world_dwm_survey,feng2025survey_dwm_survey,jia2025progressive_survey_misc,fu2024exploring_dwm_survey,zeng2024world_wm_dwm_survey,zhu2024sora_wm_dwm_survey}. Recent world-model-based planning methods for autonomous driving have shown substantial performance potential, but their computational efficiency still limits practical deployment~\cite{jia2025progressive_survey_misc}. A common consensus is that rich future imagination often incurs high computational costs. Explicit modeling approaches~\cite{zhang2025epona,zheng2024_occworld,gu202410_dome,li2024uniscene} typically generate high-dimensional images, occupancy, or explicit scene representations. Although implicit modeling approaches~\cite{gao2024_vista,li2024think2drive,zheng2025_world4drive,ni2025_maskgwm,jia2026driveworldvla} avoid some reconstruction overhead, they may still require complex state transitions and candidate evaluation. These costs further increase when future prediction extends to multiple candidate actions, creating a pronounced trade-off between planning performance and efficiency.

To address this trade-off, some works use knowledge distillation or pretrained knowledge transfer to balance performance and deployment efficiency. For example, WPT~\cite{jiang2026wpt} transfers a world-model-guided teacher policy to a lightweight student policy through policy distillation and world reward distillation. Drive-JEPA~\cite{wang2026drivejepa} combines a pretrained video encoder~\cite{assran2025vjepa2} with additional driving videos to learn predictive representations. These methods offer effective practical solutions, but their capabilities necessarily rely on strong teacher supervision, costly feature extractors, or additional training data. Efficiency gains at inference do not directly imply lower overall learning costs. Thus, these approaches fundamentally retain their dependence on expensive representations. Parameter-efficient adaptation also offers a complementary route~\cite{guo2026phaselora,kim2024openvla}.

\textbf{\textit{However, does high-quality planning necessarily depend on expensive representations?}} To re-examine this question, we selected several representative autonomous driving approaches~\cite{liao2025diffusiondrive,li2025end_wote,wang2026drivejepa} and analyzed the average quality of both their candidate trajectories and final selected trajectories across a subset of NAVSIM~\cite{dauner2024navsim}, as illustrated in Fig.~\ref{fig:motivation}(a). A key observation is that the trajectories ultimately selected by all models often fail to realize the full potential of their candidate sets. Moreover, Drive-JEPA~\cite{wang2026drivejepa} achieves similar candidate trajectory quality with feature extractors of different sizes. This suggests that representation quality alone does not determine the performance bottleneck. For autonomous driving world models, what matters more is understanding the futures associated with different candidates and making accurate selections accordingly. Therefore, improving planning performance requires not only richer future imagination but also more effective use of limited future information for candidate evaluation.

The above analysis motivates us to explore \textbf{\textit{a more direct pipeline: learning future information that is truly useful for planning under a limited computation budget via natively lightweight architectural design that constrains the cost of future state construction and trajectory evaluation}}. As shown in Fig.~\ref{fig:motivation}(b), we propose FastWorld, a lightweight world model framework for efficient autonomous driving planning. First, FastWorld constructs a compact world state that preserves camera identities and coarse spatial layouts. Second, FastWorld generates a candidate trajectory set and preliminarily filters a small subset of top candidates for low-cost visual verification, confining expensive cross-view feature interactions to limited candidates while preserving shared candidate context. Built upon this, the lightweight state transition network predicts compact future world states conditioned on the current state, ego motion, and candidate actions. Finally, a small MLP reads out future utility and corrects the initial score with bounded residuals. In this way, FastWorld allocates limited computation to both `predicting the futures' and `assessing the value of these futures for decision-making', avoiding full high-dimensional scene generation or complex future feature decoding for each candidate trajectory. Meanwhile, FastWorld does not rely on teacher distillation or extra domain-specific pre-training to achieve lightweight capacity. On the NAVSIM~\cite{dauner2024navsim,cao2025pseudo_navsimv2} benchmark, FastWorld achieves an inference speed of 132 FPS with planning performance comparable to state-of-the-art methods (92.2 PDMS and 89.4 EPDMS), see Fig.~\ref{fig:motivation}(c). On the nuScenes~\cite{NUSCENES} benchmark, FastWorld achieves a 3s average L2 of 0.40 and a 3s average CR of 0.31, at 17 FPS. The competitive inference latency demonstrates FastWorld's potential to attain a better balance between planning quality and computational efficiency.

In summary, our main contributions are as follows:
\begin{itemize}
    \setlength{\itemsep}{1pt}
    \setlength{\parskip}{0pt}
    \setlength{\topsep}{2pt}
    \item \textbf{We identify the candidate trajectory selection bottleneck in autonomous driving models.} We observe a substantial gap between the latent performance of the candidate set and the final selection in most models. Therefore, enhancing candidate understanding and evaluation is a promising direction for improving efficient planning.
    \item \textbf{We propose FastWorld, a natively lightweight world model framework.} With compact state representation, shared state transition, and selective candidate verification, FastWorld structurally controls the computational cost of future prediction and trajectory evaluation.
    \item \textbf{FastWorld achieves a competitive balance between performance and efficiency.} Without teacher distillation or extra domain-specific pre-training, FastWorld achieves 92.2 PDMS at 132 FPS on NAVSIMv1~\cite{dauner2024navsim}, 89.4 EPDMS on NAVSIMv2~\cite{cao2025pseudo_navsimv2}, and 0.40 m average L2 error at 17 FPS on nuScenes~\cite{NUSCENES}, offering an efficient world-model-based solution for real-time autonomous driving. 
\end{itemize}

\section{Related Work}
\label{sec:relatedwork}

\noindent \textbf{End-to-End Autonomous Driving.} 
End-to-end autonomous driving aims to directly map raw sensor inputs to planned trajectories. Recent research has increasingly focused on multi-modal trajectory generation~\cite{song2025diver,liu2026guideflow,song2025don_momad,zheng2025diffusionplanner_iclr2025} and closed-loop robustness~\cite{zheng2025diffusionplanner_iclr2025,renz2025simlingo}. DIVER~\cite{song2025diver} combines diffusion-based multi-modal trajectory generation with reinforcement learning. GuideFlow~\cite{liu2026guideflow} models multi-modal trajectory distributions through constraint-guided flow matching. Diffusion Planner~\cite{zheng2025diffusionplanner_iclr2025} uses a DiT to jointly model ego-vehicle planning and other-agent prediction, employing classifier guidance to generate multi-modal driving behaviors and demonstrating closed-loop robustness on nuPlan~\cite{caesar2021nuplan}. These methods improve planning capabilities in complex driving scenarios, but temporal interactions and iterative trajectory generation also pose challenges to inference efficiency~\cite{song2026momworld}. To address these challenges, existing approaches adopt strategies such as task parallelization (e.g., PARA-Drive~\cite{weng2024drive_para-drive} and SparseDrive~\cite{sun2024sparsedrive}), sparse scene representations (e.g., SSR~\cite{li2025navigation_ssr}, SparseDrive~\cite{sun2024sparsedrive}, and DriveTransformer~\cite{jia2025drivetransformer}), and fewer trajectory generation steps (e.g., DiffusionDrive~\cite{liao2025diffusiondrive} and GoalFlow~\cite{xing2025goalflow}), reducing computational overhead and supporting progress toward safe real-world deployment.

\noindent \textbf{Efficient World Modeling for Autonomous Driving.} 
World models for autonomous driving increasingly integrate future scene generation with trajectory planning. However, explicit scene generation often incurs substantial computational overhead~\cite{zhang2025epona,wang2024_drive-WM,chen2025drivinggpt}. Some studies model future scenes in latent space~\cite{jia2026driveworldvla,li2024enhancing_law,li2025end_wote,li2025drivevlaw0}. For example, DriveWorld-VLA~\cite{jia2026driveworldvla} connects world modeling and action planning through shared latent states, avoiding costly pixel-level future rollouts. Although latent-space modeling reduces visual reconstruction costs, rich imagination of future worlds may still involve multi-step state prediction and action-conditioned interactions. Other works further prioritize efficiency in world model design~\cite{xu2025delta_adtt,dang2026sparseworld,wang2026latentwam,yang2026geniedrive}. For example, SparseWorld~\cite{dang2026sparseworld} uses sparse scene representations to reduce future prediction costs. Latent-WAM~\cite{wang2026latentwam} compresses multi-view images into compact scene tokens using learnable queries. Building on compact latent-space modeling, FastWorld further reduces inference latency through compact scene representations, short-horizon future prediction, and selective trajectory refinement.

\section{Method}
\label{sec:method}

\begin{figure*}[!t]
\centering
\includegraphics[width=1.0\textwidth]{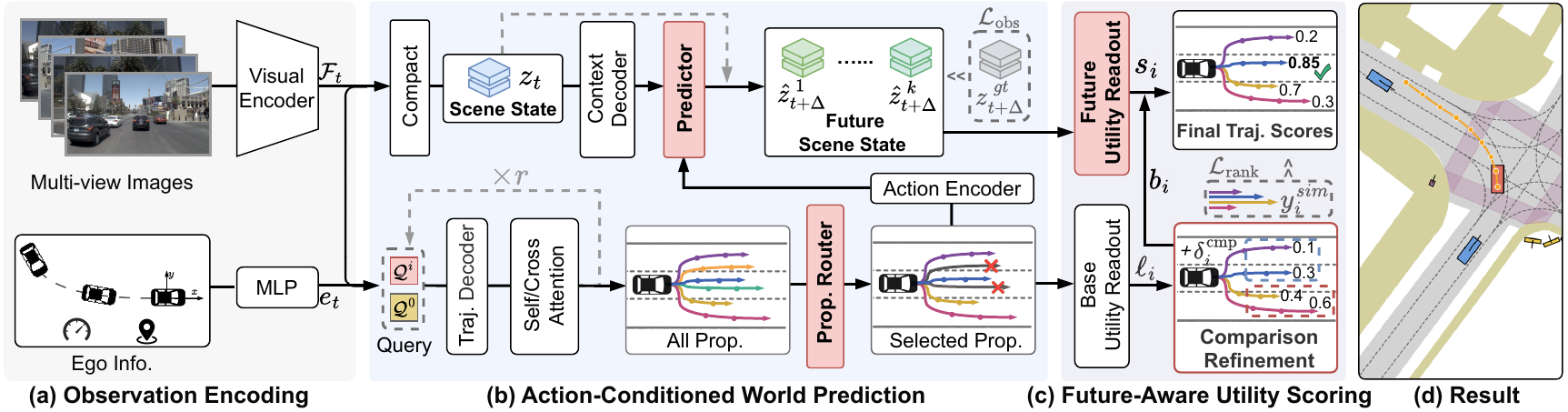}
\caption{\textbf{FastWorld Pipeline}.
FastWorld achieves efficient planning through a natively lightweight architecture.
(a) Observation Encoding extracts shared multi-view features and constructs a compact scene state.
(b) Action-Conditioned World Prediction retains a budgeted subset of trajectory proposals and predicts their short-horizon future states in parallel.
(c) Future-Aware Utility Scoring refines candidate scores using the predicted futures and selects the final trajectory. 
(d) Future-informed trajectory selection yields the final driving decision.
}
\label{fig:pipeline}
\end{figure*}

FastWorld supports the planning task through compact, action-conditioned future prediction, as shown in Fig.~\ref{fig:pipeline}. FastWorld comprises three main components: Observation Encoding (Sec.~\ref{subsec:fw_encoding}), Action-Conditioned World Prediction (Sec.~\ref{subsec:acwp}, ACWP), and Future-Aware Utility Scoring (Sec.~\ref{subsec:faus}, FAUS). The visual encoder constructs the scene state and provides trajectory-aligned evidence for candidate trajectory generation under a computational budget. ACWP predicts compact future states conditioned on the current scene and candidate actions, while FAUS uses these predictions to refine candidate trajectory scores.

\subsection{Observation Encoding}
\label{subsec:fw_encoding}
 Given $V$ synchronized camera images $\mathcal{I}_t=\{I_t^v\}_{v=1}^{V}$ and the current ego state $e_t$ with local pose, velocity, acceleration and a navigation command, FastWorld first constructs a shared visual memory $\mathcal{F}_t \in \mathbb{R}^{V \times C_{vis} \times H_{vis} \times W_{vis}}$:
\begin{equation}
\mathcal{F}_t=f_{\mathrm{vis}}(\mathcal{I}_t,e_t)
\label{eq:ft}
\end{equation}
where $f_{\mathrm{vis}}(\cdot)$ denotes the shared visual encoder with multi-scale feature fusion~\cite{he2016deep_resnet,lin2017feature_fpn,liu2021swin_transformer}. 
Next, FastWorld uses $\mathcal{F}_t$ to construct a compact scene representation $z_t \in \mathbb{R}^{V \times C_z \times H_z \times W_z}$ at time $t$:
\begin{equation}
z_t=f_{\mathrm{comp}}(\mathcal{F}_t)
\label{eq:zt}
\end{equation}
where $f_{\mathrm{comp}}(\cdot)$ applies spatial pooling within each view, a fixed orthogonal projection, and normalization. This representation preserves camera identity and coarse spatial layout. The current observation is encoded once and shared by proposal generation and world prediction.

\subsection{Action-Conditioned World Prediction}
\label{subsec:acwp}
\noindent\textbf{Trajectory proposal refinement.} 
The proposal generator conditions a bank of learnable queries $\mathcal{Q}$ on the ego state to obtain the initialized query $\mathcal{Q}^{(0)}$.
At each refinement stage $r \in \{1,\ldots,R\}$, we decode candidate trajectories from $\mathcal{Q}^{(r-1)}$ and use the predicted poses to gather visual features and update the queries to $\mathcal{Q}^{(r)}$:
\begin{equation}
\begin{aligned}
\mathcal{T}^{(r)} &= f_{\mathrm{dec}}^{(r)}(\mathcal{Q}^{(r-1)}), \\
\mathcal{Q}^{(r)} &= f_{\mathrm{align}}(
    \mathcal{Q}^{(r-1)},\mathcal{F}_t,f_{\mathrm{proj}}(\mathcal{T}^{(r)}))
\end{aligned}
\label{eq:qr}
\end{equation}
where $\mathcal{T}^{(r)}$ and $f_{\mathrm{dec}}^{(r)}(\cdot)$ denote the trajectory and its decoder for stage $r$. For queries undergoing visual verification, $f_{\mathrm{proj}}(\cdot)$ projects pose-based reference points into the camera views. The shared interaction module $f_{\mathrm{align}}(\cdot)$ combines query self-attention with cross-attention to visual memory at the projected reference locations~\cite{li2024bevformer}. 
After $R$ stages, we obtain $\mathcal{T}^{(R)}$, the final proposal set, and the corresponding query features $\mathcal{Q}^{(R)}$. Each proposal $\tau_i$ contains local positions and headings over the full planning horizon.

\noindent\textbf{Budgeted visual verification.} 
To control the final verification cost, a lightweight router uses candidate features and the shared native scorer to select a set $\mathcal{S}$ of $K$ candidate indices before the final visual update. In the final refinement stage, image cross-attention and the subsequent feed-forward computation operate only on the selected queries, while self-attention preserves the context of all candidates. We use the same indices $\mathcal{S}$ to gather the retained query features $\mathcal{Q}_{\mathcal{S}}^{(R)}$ from $\mathcal{Q}^{(R)}$ and their corresponding trajectories $\mathcal{T}_{\mathcal{S}}^{(R)}$ from $\mathcal{T}^{(R)}$.

\noindent\textbf{Action encoding.} 
We distinguish the prediction horizon $\Delta$ from the full planning horizon and encode only the trajectory prefix within $\Delta$ into an action condition $a_i$:
\begin{equation}
a_i=f_{\mathrm{act}}(\tau_i^{\leq\Delta}), \tau_i \in \mathcal{T}_{\mathcal{S}}^{(R)} 
\label{eq:ai}
\end{equation}
where $f_{\mathrm{act}}(\cdot)$ denotes the future action encoder, and $\tau_i^{\leq\Delta}$ denotes the prefix of candidate trajectory $\tau_i$ within the prediction horizon $\Delta$. This temporal alignment restricts the transition input to actions within the modeled interval.

\noindent\textbf{Parallel future prediction.} 
Next, we combine the compact scene state $z_t$ (see Eq.~\ref{eq:zt}) and the ego state $e_t$ to obtain a shared context vector $c_t$:
\begin{equation}
c_t=f_{\mathrm{ctx}}(z_t,e_t)
\label{eq:ct}
\end{equation}
where $f_{\mathrm{ctx}}(\cdot)$ denotes the context encoder. For each candidate $i$, we add the increment to the current scene state $z_t$ to obtain the predicted future state $\hat z_{t+\Delta}^{\,i}$:
\begin{equation}
\hat z_{t+\Delta}^{\,i}=z_t+f_{\mathrm{dyn}}(c_t,a_i)
\label{eq:zthat}
\end{equation}
where $f_{\mathrm{dyn}}(\cdot)$ maps the shared context and action condition to a scene-state increment. The predicted state $\hat z_{t+\Delta}^{\,i}$ has the same shape as $z_t$. All candidates share the transition parameters and are processed in parallel. The module directly predicts the target-time state without auto-regressive rollout or pixel reconstruction.

\subsection{Future-Aware Utility Scoring}
\label{subsec:faus}
\noindent\textbf{Pairwise comparison prior.} 
For each retained candidate, FAUS first applies temporal max pooling to its query features and passes the pooled feature to the shared native scorer to obtain a base utility score $\ell_i$. It then ranks the candidates by these scores and pairs candidates at corresponding ranks in the upper and lower halves of the ranked list. We refine $\ell_i$ through pairwise comparison to obtain the comparison prior $b_i$:
\begin{equation}
b_{i \in \mathcal{S}}=\ell_i+\delta_i^{\mathrm{cmp}}
\label{eq:bi}
\end{equation}
where $\delta_i^{\mathrm{cmp}}$ is a bounded correction computed by the comparison module. The module compares paired candidates using their retained query features $\mathcal{Q}_{\mathcal{S}}^{(R)}$ and trajectories $\mathcal{T}_{\mathcal{S}}^{(R)}$ (see Sec.~\ref{subsec:acwp}), along with the current ego state $e_t$ and the base utility scores. To compute the correction for candidate $i$, it combines information from the pair containing that candidate with context aggregated across all pairs.

\noindent\textbf{Future-aware reranking.} 
For each retained candidate, FAUS uses the predicted future state $\hat z_{t+\Delta}^{\,i}$ (see Eq.~\ref{eq:zthat}) to refine the comparison prior $b_i$ and obtain the final utility score $s_i$:
\begin{equation}
s_{i \in \mathcal{S}}=b_i+f_{\mathrm{util}}\bigl(\hat z_{t+\Delta}^{\,i}\bigr)
\label{eq:si}
\end{equation}
where $f_{\mathrm{util}}(\cdot)$ is a lightweight utility readout that maps the predicted state to a scalar correction with magnitude bounded by $\rho_u$. 
With this, FastWorld efficiently selects the highest-scoring retained trajectory through shared visual encoding, budgeted visual verification, and lightweight scoring of compact future states predicted in parallel.

\subsection{Training Objective}
\label{subsec:supervision}

\noindent\textbf{Future state prediction supervision.}
For the logged action, we match the predicted future state $\hat z_{t+\Delta}^{\mathrm{log}}$ (see Eq.~\ref{eq:zthat}) to the observed future state $z_{t+\Delta}^{\star}$ (see Eq.~\ref{eq:zt}), yielding the observation loss $\mathcal{L}_{\mathrm{obs}}$:
\begin{equation}
\mathcal{L}_{\mathrm{obs}}=\mathbb{E}_{\mathrm{valid}}\!\left[D_{\mathrm{obs}}\bigl(\hat z_{t+\Delta}^{\mathrm{log}},z_{t+\Delta}^{\star}\bigr)
\right]
\label{eq:loss_obs}
\end{equation}
where $D_{\mathrm{obs}}(\cdot)$ combines smooth L1 loss with the mean tokenwise cosine distance and $\mathbb{E}_{\mathrm{valid}}(\cdot)$ denotes the expectation over samples with valid future observations.

\begin{table*}[!t]
\centering
\captionsetup{skip=2pt}
\setlength{\intextsep}{2pt}
\caption{
\textbf{Comparison with state-of-the-art methods in \textbf{\texttt{NAVSIMv1}}~\cite{dauner2024navsim}.} The \textbf{BEST} and the \underline{SECOND BEST} are indicated with bold and underline respectively. $\star$ results are cited from \cite{wang2026meanfuser}. $\dagger$ results are cited from \cite{jia2026driveworldvla}. $\ddagger$ results are reproduced using the official checkpoint. Abbreviations: Num.$\times$C: surround Num.-view cameras; L: LiDAR; NC: No Collision; DAC: Drivable Area Compliance; TTC: Time-To-Collision; EP: Ego Progress; C: Comfort; PDMS: Predictive Driver Model Score; FPS{\tiny{[GPU]}}: Frames Per Second (includes the complete forward pass and evaluated with a single [GPU]).}
\renewcommand\arraystretch{0.5}
\tabcolsep=0.8mm 
\setlength{\tabcolsep}{2.5mm}
\footnotesize
\resizebox{\linewidth}{!}{
\begin{tabular}{ll|cc|cccccc|c}
\toprule
Methods & Venue & Sensors &Backbone& NC$\uparrow$ & DAC$\uparrow$ & TTC$\uparrow$ & C$\uparrow$ & EP$\uparrow$ & PDMS$\uparrow$ & FPS$\uparrow$\tiny{[GPU]} \\
\midrule
\midrule
Human & - & - & - & 100.0 & 100.0 & 100.0 & 99.9 & 87.5 & 94.8 & - \\
\midrule
\multicolumn{11}{c}{\textit{\textcolor{gray!70}{E2E-based Methods}}}\\
TransFuser~\cite{chitta2022transfuser} & TPAMI 2022 & 3$\times$C \& L & ResNet-34 & 97.7 & 92.8 & 92.8 & \textbf{100.0} & 79.2 & 84.0 & \textsuperscript{$\star$}\underline{63} {\tiny{[H20]}} \\
Hydra-MDP~\cite{li2024hydra} & arXiv 2024 & 3$\times$C \& L & ResNet-34 & 98.3 & 96.0 & 94.6 & \textbf{100.0} & 78.7 & 86.5 & \textsuperscript{$\star$}20 {\tiny{[H20]}} \\
DiffusionDrive~\cite{liao2025diffusiondrive} & CVPR 2025 & 3$\times$C \& L & ResNet-34 & 98.2 & 96.2 & 94.7 & \textbf{100.0} & 82.2 & 88.1 & \textsuperscript{$\star$}39 {\tiny{[H20]}} \\
GoalFlow~\cite{xing2025goalflow} & CVPR 2025 & 3$\times$C \& L & V2-99 & 98.4 & \underline{98.3} & 94.6 & \textbf{100.0} & 85.0 & 90.3 & \textsuperscript{$\star$}10 {\tiny{[H20]}} \\
MeanFuser~\cite{wang2026meanfuser} & CVPR 2026 & 3$\times$C & ResNet-34 & 98.6 & 97.0 & 95.0 & \textbf{100.0} & 82.8 & 89.0 & 59 {\tiny{[H20]}} \\
\midrule
\multicolumn{11}{c}{\textit{\textcolor{gray!70}{World-Model-based Methods}}}\\
WoTE~\cite{li2025end_wote} & ICCV 2025 & 3$\times$C \& L & ResNet-34 & 98.5 & 96.8 & 94.9 & \underline{99.9} & 81.9 & 88.3 & \textsuperscript{$\ddagger$}50 {\tiny{[4090]}} \\
Epona~\cite{zhang2025epona} & ICCV 2025 & 1$\times$C & DCAE & 97.9 & 95.1 & 93.8 & \underline{99.9} & 80.4 & 86.2 & \textsuperscript{$\dagger$}0.4 {\tiny{[H20]}} \\
DriveLaW~\cite{xia2026drivelaw} & CVPR 2026 & 1$\times$C & LTX-Video 3D VAE & 99.0 & 97.1 & \underline{96.7} & \textbf{100.0} & 81.3 & 89.1 & - {\tiny{[4090]}} \\
DriveVLA-W0~\cite{li2025drivevlaw0} & ICLR 2026 & 1$\times$C & - & 98.7 & \textbf{99.1} & 95.3 & 99.3 & 83.3 & 90.2 & 13 {\tiny{[H200]}} \\
DriveVA~\cite{liu2026driveva} & ECCV 2026 & 1$\times$C & Wan2.2 3D VAE & \textbf{99.2} & 97.5 & \textbf{98.7} & \textbf{100.0} & 83.5 & 90.9 & - {\tiny{[H20]}} \\
DriveWorld-VLA~\cite{jia2026driveworldvla} & ICML 2026 & 3$\times$C & ResNet-34 & \underline{99.1} & 98.2 & 96.1 & \textbf{100.0} & 85.9 & 91.3 & 1.2 {\tiny{[H20]}} \\
MomWorld~\cite{song2026momworld} & arXiv 2026 & 3$\times$C \& L & V2-99 & 98.6& 98.2& 93.9& \textbf{100.0}& 85.7& 90.2 & 7.2 {\tiny{[4090]}} \\
Drive-JEPA~\cite{wang2026drivejepa} & arXiv 2026 & 1$\times$C & ViT-L & \underline{99.1} & 98.2 & 96.2 & \underline{99.9} & \textbf{91.4} & \textbf{93.7} & \textsuperscript{$\ddagger$}18 {\tiny{[4090]}} \\
DriveFuture~\cite{hong2026drivefuture} & NeurIPS 2026 & 6$\times$C & V2-99 & 98.8 & \textbf{99.1} & 95.4 & \textbf{100.0} & 84.2 & 90.7 & - {\tiny{[5090]}} \\
\rowcolor{cyan!15} \textbf{FastWorld (Ours)} & - & 4$\times$C & ResNet-34 & 98.5 & 98.0 & 94.6 & \underline{99.9} & \underline{89.9} & \underline{92.2} & \textbf{132} {\tiny{[4090]}} \\
\bottomrule
\end{tabular}
}
\label{tab:main_navsimv1}
\end{table*}

\noindent\textbf{Decision supervision.}
We use simulator-derived utilities $y_i$ to supervise the final candidate scores $s_i$ (see Eq.~\ref{eq:si}). To rank candidates with higher utility above those with lower utility, we use the pairwise loss $\mathcal{L}_{\mathrm{rank}}$:
\begin{equation}
\mathcal{L}_{\mathrm{rank}}
=\sum_{(m,n)\in\mathcal{P}} w_{mn} \times \operatorname{SoftPlus}\!\left(\frac{s_n-s_m}{T_s}\right)
\label{eq:loss_rank}
\end{equation}
where $\mathcal{P}$ contains ordered pairs $(m,n)$ drawn from $\mathcal{E}$ that satisfy $y_m-y_n>\eta$. The set $\mathcal{E}$ contains retained candidates whose comparison priors lie within $\varepsilon$ of the highest prior. The weight $w_{mn}$ is the simulator utility gap $y_m-y_n$ normalized over $\mathcal{P}$, and $T_s$ is the score temperature. The $\operatorname{SoftPlus}(\cdot)$ penalty increases as the score $s_n$ of the lower-utility trajectory approaches or exceeds the score $s_m$ of the higher-utility trajectory. 
We also encourage the model to assign more probability to candidates with higher simulator utility through $\mathcal{L}_{\mathrm{util}}$:
\begin{equation}
\mathcal{L}_{\mathrm{util}}
=u_{\max}-\sum_{i\in\mathcal{E}}p_i y_i
\label{eq:loss_utility}
\end{equation}
where $u_{\max}$ is the utility upper bound, and $p_i$ is the softmax probability of $s_i$ over $\mathcal{E}$ at temperature $T_s$. We average both decision losses over scenes with valid pairs.

\noindent\textbf{Total training objective.}
We combine the three supervision terms and the regularizers into the total training objective $\mathcal{L}$:
\begin{equation}
\mathcal{L}
=\lambda_{\mathrm{obs}}\mathcal{L}_{\mathrm{obs}}
+\lambda_{\mathrm{rank}}\mathcal{L}_{\mathrm{rank}}
+\lambda_{\mathrm{util}}\mathcal{L}_{\mathrm{util}}
+\mathcal{L}_{\mathrm{reg}}
\label{eq:loss_total}
\end{equation}
where $\mathcal{L}_{\mathrm{reg}}$ is the weighted sum of the residual, imitation, and planner consistency losses (detailed in the Supplementary Material). And the $\lambda$ control their contributions.

\section{Experiments}
\label{sec:experiments}

\subsection{Details}

\noindent \textbf{Dataset.} We evaluate FastWorld on \textbf{\texttt{NAVSIMv1}}~\cite{dauner2024navsim}, \textbf{\texttt{NAVSIMv2}}~\cite{cao2025pseudo_navsimv2} and \textbf{\texttt{nuScenes}}~\cite{NUSCENES}. 
\textbf{\texttt{NAVSIMv1}} is an autonomous driving planning benchmark built on the OpenScene~\cite{openscene2023_official} dataset, providing 120 hours of driving data with multi-camera inputs and fused LiDAR point clouds. 
The core metric Predictive Driver Model Score (PDMS) is defined as the product of a penalty term and a weighted average score. 
\textbf{\texttt{NAVSIMv2}} proposes a two-stage pseudo-simulation evaluation. Its core metric is called the Extended Predictive Driver Model Score (EPDMS). 
For open-loop planning, \textbf{\texttt{nuScenes}} is a large-scale outdoor driving dataset with 1000 multi-modal scenes. Each scene lasts 20 s, is annotated at 2 Hz, and includes 6 synchronized camera images and LiDAR point clouds. We use L2 and Collision Rate (CR) as evaluation metrics. 
See the Supplementary Material for more information about the datasets.

\noindent \textbf{Implementation.}
For \textbf{\texttt{NAVSIM}}, we take synchronized images from four cameras (\texttt{CAM\_F0}, \texttt{CAM\_B0}, \texttt{CAM\_L0}, \texttt{CAM\_R0}) as input. A shared ResNet-34~\cite{he2016deep_resnet} with FPN~\cite{lin2017feature_fpn} outputs a $14\times24$, 256-channel feature map per view. For world prediction, each view is pooled to a $2\times2$ grid and projected to 64 channels, compressing visual memory from 1,344 spatial tokens to 16 scene tokens. The planner generates 64 trajectories; we keep 16 candidates for visual verification and predict a compact 1-second future state for each.
Training uses AdamW with a global batch size of 128. The model for Tab.~\ref{tab:main_navsimv1} and Tab.~\ref{tab:main_navsimv2} is trained for 30 epochs on 8 RTX 4090 GPUs.
For \textbf{\texttt{nuScenes}}, we use six surround-view cameras. A Swin-T~\cite{liu2021swin_transformer} encoder processes the current frame and a frame from roughly 1 second prior. The utility readout employs 36 tokens (width 64) per prediction step to rerank 16 out of 64 proposals over a 3-second horizon. The model for Tab.~\ref{tab:main_nuscenes} is trained for 95 epochs on the same hardware.
See the Supplementary Material for additional implementation details.

\subsection{Main Results}

\begin{table*}[!t]
\centering
\captionsetup{skip=2pt}
\setlength{\intextsep}{2pt}
\caption{
\textbf{Comparison with state-of-the-art methods in \textbf{\texttt{NAVSIMv2}}~\cite{cao2025pseudo_navsimv2}.} The \textbf{BEST} and the \underline{SECOND BEST} are indicated with bold and underline respectively. $\ddagger$ results are reproduced using the official checkpoint. Abbreviations: NC: No at-fault Collision; DAC: Drivable Area Compliance; DDC: Driving Direction Compliance; TLC: Traffic Light Compliance; EP: Ego Progress; TTC: Time to Collision; LK: Lane Keeping; HC: History Comfort; EC: Extended Comfort; EPDMS: Extended Predictive Driver Model Score.}
\renewcommand\arraystretch{0.5}
\setlength{\tabcolsep}{2.2mm}
\footnotesize
\resizebox{\linewidth}{!}{
\begin{tabular}{c|ll|c|ccccccccccc}
\toprule
Split&Methods &Venue&Backbone&Stage& NC$\uparrow$ & DAC$\uparrow$ & DDC$\uparrow$ & TLC$\uparrow$ & EP$\uparrow$ & TTC$\uparrow$ & LK$\uparrow$ & HC$\uparrow$ & EC$\uparrow$ & EPDMS$\uparrow$ \\
\midrule
\midrule
\multirow{20.5}{*}{\rotatebox{90}{\textbf{\texttt{navtest}}}}& \multicolumn{14}{c}{\textit{\textcolor{gray!70}{E2E-based Methods}}}\\
&TransFuser~\cite{chitta2022transfuser} &TPAMI 2022&ResNet-34&-& 96.9 & 89.9 & 97.8 & 99.7 & 87.1 & 95.4 & 92.7 & \textbf{98.3} & 87.2 & 76.7 \\
&HydraMDP++~\cite{li2025hydra_mdp++}   &arXiv 2025&V2-99&-&\textbf{98.8} & 97.8 & 99.1 & \textbf{100.0} & 84.0 & 95.3 & 70.1 & - & \textbf{96.8} & 84.1\\
&DiffusionDrive~\cite{liao2025diffusiondrive}  &CVPR 2025&ResNet-34&-& 98.2 &  95.9 & 99.4 & 99.8 & 87.5 & 97.3 & 96.8 & \textbf{98.3} & 87.7 & 84.5\\
&MeanFuser~\cite{wang2026meanfuser}  &CVPR 2026&ResNet-34&-& 98.3& 97.2& \textbf{99.6}& 99.8& 87.6& 97.4& \underline{97.3}& \textbf{98.3}& 88.2& 89.5\\
\cmidrule(lr){2-15}
&  \multicolumn{14}{c}{\textit{\textcolor{gray!70}{World-Model-based Methods}}}\\
&DriveVLA-W0~\cite{li2025drivevlaw0} &ICLR 2026&-&-&98.5& \textbf{99.1}& 98.0& 99.7& 86.4& 98.1& 93.2& 97.9& 58.9& 86.1 \\
&Latent-WAM~\cite{wang2026latentwam}&ACMMM 2026 &DINOv2-B&-&98.1& 97.3& \textbf{99.6}& 99.8 &87.7& 97.3& \textbf{97.6}& \underline{98.1}& 87.3& 89.3\\
&DriveWorld-VLA~\cite{jia2026driveworldvla}&ICML 2026&ResNet-34&-& \underline{98.6} & \textbf{99.1} & \textbf{99.6} & 99.8 & 87.4 & 97.9 & 97.0 & 97.8 & 78.6 & 86.8 \\
&MomWorld~\cite{song2026momworld} &arXiv 2026&V2-99&-& 98.1& 98.1& \underline{99.5}& 99.8& \underline{88.9}& \underline{98.2}& 96.1& \textbf{98.3}& \underline{90.6}& \textbf{90.1}\\
&Drive-JEPA~\cite{wang2026drivejepa} &arXiv 2026&ViT-L&-& 98.4 & \underline{98.6} & 99.1 & 99.8 & 88.4 & 97.8 & \textbf{97.6} & 97.9 & 84.8 & 87.8\\
&DriveFuture~\cite{hong2026drivefuture} &NeurIPS 2026&V2-99&-& \textbf{98.8}& \textbf{99.1}& \textbf{99.6}& \underline{99.9}& 86.6& \textbf{98.4}& 96.4& \textbf{98.3}& 74.8& \underline{89.9} \\
\rowcolor{cyan!15} \cellcolor{white}& \textbf{FastWorld (Ours)} &-&ResNet-34&-& 98.2 & 97.4 & \underline{99.5} & 99.6 & \textbf{89.3} & 97.5 & 96.8 & \textbf{98.3} & 86.2 & 89.4 \\
\midrule
\midrule
\multirow{24}{*}{\rotatebox{90}{\textbf{\texttt{navhard}}}}&  \multicolumn{14}{c}{\textit{\textcolor{gray!70}{E2E-based Methods}}}\\
&\multirow{2}{*}{\textsuperscript{$\ddagger$}DiffusionDrive~\cite{liao2025diffusiondrive}}&\multirow{2}{*}{CVPR 2025}&\multirow{2}{*}{ResNet-34}&1&96.0& 79.7& 97.4& 99.5& 81.3& 93.1& 90.8 &96.8& 73.8&\multirow{2}{*}{24.2} \\
&&&&2&82.1& 72.2& 88.5& 98.7& 85.1& 78.8& 49.2& 89.3& 71.2& \\
\cmidrule(lr){5-15}
&\multirow{2}{*}{GuideFlow~\cite{liu2026guideflow}}&\multirow{2}{*}{CVPR 2026}&\multirow{2}{*}{ResNet-34}&1&97.8& 97.1& 100.0& 100.0& 81.4& 98.5& 91.4& 92.8& 34.2&\multirow{2}{*}{\underline{46.7}} \\
&&&&2&87.3& 92.3& 98.0& 96.9& 75.8& 85.5& 59.3& 95.4& 53.5& \\
\cmidrule(lr){2-15}
&  \multicolumn{14}{c}{\textit{\textcolor{gray!70}{World-Model-based Methods}}}\\
&\multirow{2}{*}{MomWorld~\cite{song2026momworld}}&\multirow{2}{*}{arXiv 2026}&\multirow{2}{*}{V2-99}&1& 96.9& 93.6& 99.8& 99.8& 80.4& 96.9& 96.4& 97.6& 60.0&\multirow{2}{*}{42.8} \\
&&&&2& 86.4& 88.2& 94.2& 98.3& 81.7& 84.4& 55.3& 97.0& 54.7& \\
\cmidrule(lr){5-15}
&\multirow{2}{*}{DriveFuture~\cite{hong2026drivefuture}}&\multirow{2}{*}{NeurIPS 2026}&\multirow{2}{*}{V2-99}&1&99.8& 99.8& 100.0& 99.6& 85.7& 99.8& 98.7& 97.6& 66.2&\multirow{2}{*}{\textbf{55.5}} \\
&&&&2&90.6& 87.5& 94.1& 99.1& 84.6& 88.8& 58.3& 93.5& 45.6& \\
\cmidrule(lr){5-15}
&\multirow{2}{*}{\textsuperscript{$\ddagger$}Drive-JEPA~\cite{wang2026drivejepa}}&\multirow{2}{*}{arXiv 2026}&\multirow{2}{*}{ViT-L}&1&95.3& 96.4& 99.4& 99.8& 85.1& 94.9& 96.9 &97.3& 35.1&\multirow{2}{*}{37.7} \\
&&&&2&87.8& 86.0& 92.3 &99.1 &80.2& 84.7 &54.9 &93.0& 27.2& \\
\cmidrule(lr){5-15}
\rowcolor{cyan!15} \cellcolor{white}& &&&1&96.7 & 92.9 & 99.0 & 99.3 & 80.0 & 96.7 & 92.2 & 97.3 & 60.0& \\
\rowcolor{cyan!15} \cellcolor{white}& \multirow{-2}{*}{\textbf{FastWorld (Ours)}}& \multirow{-2}{*}{-}&\multirow{-2}{*}{ResNet-34}&2&88.0 & 86.7 & 94.2 & 98.5 & 77.8 & 85.0 & 61.0 & 95.4 & 51.9&\multirow{-2}{*}{43.9} \\
\bottomrule
\end{tabular}
}
\label{tab:main_navsimv2}
\end{table*}

\begin{table*}[!t]
\centering
\captionsetup{skip=2pt} 
\setlength{\intextsep}{2pt}
\caption{
\textbf{Comparison with state-of-the-art methods in \texttt{\textbf{nuScenes}}~\cite{NUSCENES} validation set.} The \textbf{BEST} and the \underline{SECOND BEST} are indicated with bold and underline respectively. $\dagger$ results are cited from \cite{jia2026driveworldvla}. $\ddagger$ results are reproduced using the official checkpoint. For a fair comparison, FastWorld makes no use of ego status information. Abbreviations: C: surround 6-view cameras ($\ast$ indicates the use of only the front camera), CR: Collision Rate; FPS{\tiny{[GPU]}}: Frames Per Second (includes the complete forward pass and evaluated with a single [GPU]).}
\renewcommand\arraystretch{0.5}
\tabcolsep=0.8mm 
\setlength{\tabcolsep}{2.3mm}
\footnotesize
\resizebox{\linewidth}{!}{
\begin{tabular}{ll|cc|cccc|cccc|c}
\toprule
\multirow{2.5}{*}{Methods}&\multirow{2.5}{*}{Venue}&\multirow{2.5}{*}{Sensors}&\multirow{2.5}{*}{Backbone}&\multicolumn{4}{c|}{L2(m)$\downarrow$}&\multicolumn{4}{c|}{CR(\%)$\downarrow$}&\multirow{2.5}{*}{FPS$\uparrow$\tiny{[GPU]}} \\
\cmidrule{5-12}
&&&&1s&2s&3s&Avg.&1s&2s&3s&Avg.&\\
\midrule
\midrule
\multicolumn{13}{c}{\textit{\textcolor{gray!70}{E2E-based Methods}}}\\
UniAD~\cite{hu2023planning_uniad} &CVPR 2023&C&ResNet-101&0.48& 0.96& 1.65& 1.03& 0.05& 0.17& 0.71& 0.31&1.8 {\tiny{[A100]}}\\
MomAD~\cite{song2025don_momad}&CVPR 2025&C&ResNet-50&0.43& 0.88& 1.62& 0.98& 0.06& \underline{0.16}& 0.68& 0.30&7.8 {\tiny{[4090]}}\\
\midrule
\multicolumn{13}{c}{\textit{\textcolor{gray!70}{World-Model-based Methods}}}\\
GenAD~\cite{zheng2024_ECCV_genad}&ECCV 2024&C &ResNet-50& 0.36 & 0.83 & 1.55 & 0.91 & 0.06 & 0.23 & 1.00 & 0.43& 6.7 {\tiny{[3090]}}\\
\textsuperscript{$\ddagger$}LAW~\cite{li2024enhancing_law} & ICLR 2025 &  C   &Swin-T&  0.31& 0.61& 1.02& 0.65&0.27&0.21&0.54&0.34&\textsuperscript{$\ddagger$}\textbf{19} {\tiny{[4090]}} \\
Epona~\cite{zhang2025epona}&ICCV 2025&\textsuperscript{$\ast$}C&DCAE& 0.61& 1.17& 1.98& 1.25& \underline{0.01}& 0.22& 0.85& 0.36&\textsuperscript{$\dagger$}0.4 {\tiny{[H20]}}\\
DriveWorld-VLA~\cite{jia2026driveworldvla}   & ICML 2026 &  C & Swin-T&0.28&0.58&0.99&0.61&\textbf{0.00}&\textbf{0.10}&\textbf{0.38}&\textbf{0.16}&1.2 {\tiny{[H20]}}\\
MomWorld~\cite{song2026momworld}&arXiv 2026&C&ResNet-50&\underline{0.27}& \underline{0.52}& \underline{0.86}&\underline{0.55}& 0.02& 0.17& \underline{0.42}& \underline{0.20}&7.2 {\tiny{[4090]}}\\
\rowcolor{cyan!15} \textbf{FastWorld (Ours)} & - & C & Swin-T & \textbf{0.21} & \textbf{0.37} & \textbf{0.63} & \textbf{0.40} & 0.10 & 0.30 & 0.52 & 0.31 & \underline{17} {\tiny{[4090]}}\\
\bottomrule
\end{tabular}
}
\label{tab:main_nuscenes}
\end{table*}

\noindent \textbf{\texttt{NAVSIMv1}.}
As shown in Tab.~\ref{tab:main_navsimv1}, FastWorld achieves 92.2 PDMS at 132 FPS on a single RTX 4090, surpassing existing approaches by a large margin in FPS while maintaining competitive PDMS. Only Drive-JEPA~\cite{wang2026drivejepa} boasts a modest PDMS advantage among competing planners (92.2 vs. 93.7). Running on the same GPU, FastWorld reaches 7.3$\times$ its throughput (132 vs. 18), at the cost of just -1.5 PDMS. FastWorld also exceeds DriveWorld-VLA~\cite{jia2026driveworldvla} by 0.9 PDMS points (92.2 vs. 91.3) and achieves the second-highest EP (89.9). These results show that FastWorld can run substantially faster while remaining close to the best planning score in the comparison.

\noindent \textbf{\texttt{NAVSIMv2}.}
Tab.~\ref{tab:main_navsimv2} presents the comparison with state-of-the-art methods on the \textbf{\texttt{navtest}} and the two-stage \textbf{\texttt{navhard}}. On the \textbf{\texttt{navtest}} split, FastWorld achieves a competitive EPDMS of 89.4, outperforming most existing E2E and World-Model-based methods. Although MomWorld~\cite{song2026momworld} achieves the highest EPDMS of 90.1, its inference speed is limited to only 7.2 FPS, and its performance on \textbf{\texttt{NAVSIMv1}} is inferior to FastWorld (90.2 vs. 92.2 in Tab.~\ref{tab:main_navsimv1}). Similarly, DriveFuture~\cite{hong2026drivefuture} achieves a slightly higher EPDMS of 89.9 but falls behind FastWorld on \textbf{\texttt{NAVSIMv1}} (90.7 vs. 92.2 in Tab.~\ref{tab:main_navsimv1}). Latent-WAM~\cite{wang2026latentwam} achieves a comparable EPDMS of 89.3 but reports an inference latency of approximately 0.11s (9 FPS) on an NVIDIA A100 GPU. On the more challenging \textbf{\texttt{navhard}} split, FastWorld achieves an EPDMS of 43.9, outperforming MomWorld~\cite{song2026momworld} (42.8) while remaining competitive with other approaches. The performance gap compared with the strongest methods could potentially be explained by our constrained computational budget, which may restrict the modeling capacity for complex, long-horizon interactions.

\begin{figure*}[!t]
\centering
\includegraphics[width=1.0\linewidth]{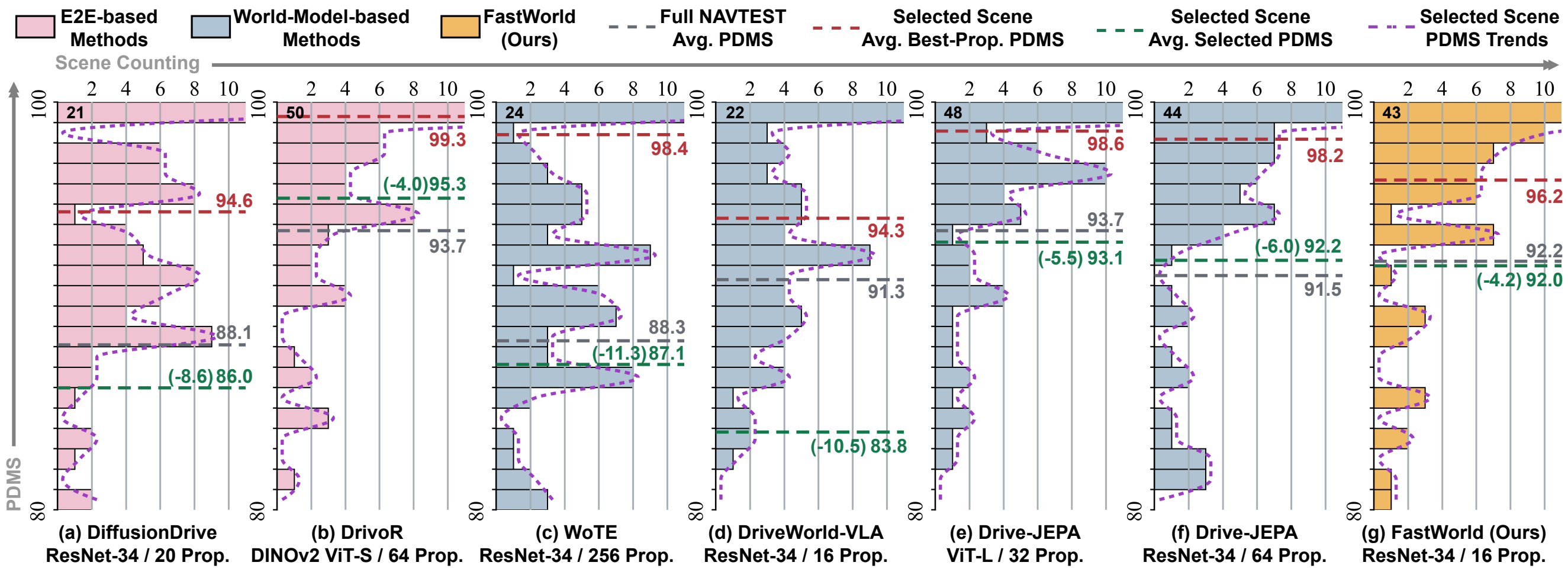}
\caption{
\textbf{Analysis of proposal potential and selection effectiveness in 100 selected \texttt{navtest} scenes.} We compare (a) DiffusionDrive~\cite{liao2025diffusiondrive}, (b) DrivoR~\cite{kirby2026driving_DrivoR}, (c) WoTE~\cite{li2025end_wote}, (d) DriveWorld-VLA~\cite{jia2026driveworldvla}, (e) \& (f) Drive-JEPA~\cite{wang2026drivejepa}, and (g) our FastWorld.
Prior methods exhibit a substantial candidate potential-selection gap. Among World-Model-based approaches, FastWorld achieves the smallest selection gap (4.2 PDMS) with limited computational overhead.
}
\label{fig:qianli}
\end{figure*}

\noindent \textbf{\texttt{nuScenes}.}
Tab.~\ref{tab:main_nuscenes} presents the quantitative comparison with state-of-the-art methods on the \textbf{\texttt{nuScenes}} validation set. The inference speed of FastWorld is measured using the dedicated \textbf{\texttt{nuScenes}} pipeline without any ego status information. FastWorld achieves state-of-the-art performance, obtaining the lowest L2 across all prediction horizons (0.21 m, 0.37 m, and 0.63 m). In terms of inference efficiency, FastWorld achieves 17 FPS, ranking second among all compared methods and closely approaching LAW~\cite{li2024enhancing_law} (19 FPS). LAW~\cite{li2024enhancing_law} benefits from its lightweight perception-free architecture. Despite this, FastWorld achieves substantially higher planning accuracy with only a marginal reduction in inference speed.

\begin{table}[!t]
\centering
\captionsetup{skip=2pt} 
\setlength{\intextsep}{2pt}
\caption{
\textbf{Analysis of model parameters and two types of FPS in \textbf{\texttt{NAVSIMv1}}~\cite{dauner2024navsim}.} The \textbf{BEST} and the \underline{SECOND BEST} are indicated with bold and underline respectively. $\star$ results are cited from \cite{wang2026meanfuser}. $\ddagger$ results reproduced with the official checkpoint. `Plan FPS' excludes the perception encoder, while `Forward FPS' includes the complete forward pass. `GPU' denotes only the device for FPS measurement, not necessarily for PDMS.}
\renewcommand\arraystretch{0.5}
\setlength{\tabcolsep}{1.5mm}
\footnotesize
\resizebox{\linewidth}{!}{
\begin{tabular}{l|c|c|cc|c}
\toprule
\multirow{2.5}{*}{Methods} & Param.$\downarrow$ & \multirow{2.5}{*}{PDMS$\uparrow$} & \multicolumn{2}{c|}{FPS$\uparrow$} & \multirow{2.5}{*}{GPU}\\
\cmidrule(lr){4-5}
&(M)&&Plan&Forward&\\
\midrule
\midrule
TransFuser~\cite{chitta2022transfuser}&  \textsuperscript{$\star$}55.9 & 84.0 & \textsuperscript{$\star$}\textbf{3934} & \textsuperscript{$\star$}\underline{63}&H20 \\
GoalFlow~\cite{xing2025goalflow}&  \textsuperscript{$\star$}62.3 & 90.3 & \textsuperscript{$\star$}11 & \textsuperscript{$\star$}10&H20 \\
Hydra-MDP~\cite{li2024hydra}&  \textsuperscript{$\star$}65.2 & 86.5 & \textsuperscript{$\star$}25 & \textsuperscript{$\star$}20&H20 \\
DiffusionDrive~\cite{liao2025diffusiondrive}&  \textsuperscript{$\star$}60.7 & 88.1 & \textsuperscript{$\star$}75 & \textsuperscript{$\star$}39&H20 \\
MeanFuser~\cite{wang2026meanfuser}&  54.6 & 89.0 & 434 & 59&H20 \\
DrivoR~\cite{kirby2026driving_DrivoR}& \textsuperscript{$\ddagger$}\underline{40.8} & \textbf{93.7} & \textsuperscript{$\ddagger$}173 & \textsuperscript{$\ddagger$}11&4090 \\
WoTE~\cite{li2025end_wote}&\textsuperscript{$\ddagger$}63.9&88.3&\textsuperscript{$\ddagger$}90&\textsuperscript{$\ddagger$}50&4090\\
Drive-JEPA~\cite{wang2026drivejepa}& \textsuperscript{$\ddagger$}311.2 & \textbf{93.7} &\textsuperscript{$\ddagger$}94 &\textsuperscript{$\ddagger$}18&4090 \\
\rowcolor{cyan!15} \textbf{FastWorld (Ours)}&  \textbf{27.0} & \underline{92.2} & \underline{520} & \textbf{132}&4090 \\
\bottomrule
\end{tabular}
}
\label{tab:plan_fps}
\end{table}

\subsection{Performance-Efficiency Analysis}

\noindent \textbf{Proposal Potential and Selection Effectiveness.}
To further investigate the trajectory selection bottleneck identified in Fig.~\ref{fig:motivation}, we analyze the quality of candidate and selected trajectories across representative planning methods. As shown in Fig.~\ref{fig:qianli}, we sample 100 scenes from the \textbf{\texttt{navtest}} split and report the average PDMS of the best candidate trajectory (red dashed line) and the finally selected trajectory (green dashed line) for each method, together with the distribution of selected trajectory scores (purple dashed line from bar).

A key observation is that high-quality trajectory generation does not necessarily translate into effective trajectory selection. Despite generating promising candidates, most methods exhibit substantial gaps between their best achievable and actual selected scores, reaching 11.3, 10.5, and 8.6 PDMS points for WoTE~\cite{li2025end_wote}, DriveWorld-VLA~\cite{jia2026driveworldvla}, and DiffusionDrive~\cite{liao2025diffusiondrive}, respectively. Moreover, the selected trajectory scores of DiffusionDrive~\cite{liao2025diffusiondrive} and WoTE~\cite{li2025end_wote} are relatively dispersed, with only 21 and 24 scenes falling into the highest PDMS interval (99–100), respectively. In contrast, several other methods exhibit greater concentrations of high scores.

Notably, FastWorld achieves a remarkably small selection gap of only 4.2 PDMS (96.2 vs. 92.0), the smallest among world-model-based methods and close to the overall minimum of 4.0 PDMS (99.3 vs. 95.3) attained by DrivoR~\cite{kirby2026driving_DrivoR}. These results empirically support our core premise: effectively leveraging limited future-related information can be more valuable than simply increasing the complexity of imagined futures.

\noindent \textbf{Model parameters and two types of FPS.}
Tab.~\ref{tab:plan_fps} compares model complexity, planning performance, and inference efficiency on \textbf{\texttt{NAVSIMv1}}. Following MeanFuser~\cite{wang2026meanfuser}, we report Plan FPS (excluding the perception encoder) and Forward FPS (full forward pass). FastWorld achieves the smallest model size at 27.0M parameters. Notably, TransFuser~\cite{chitta2022transfuser} shows a large gap between Plan FPS and Forward FPS (3934 vs. 63), indicating its latency is dominated by backbone feature extraction. In contrast, GoalFlow~\cite{xing2025goalflow} yields nearly matching Plan and Forward FPS (11 vs. 10), pointing to bottlenecks in downstream feature processing and planning. Instead of leaning toward either extreme, FastWorld uses a balanced lightweight design cutting overhead in perception and planning, enabling efficient end-to-end inference with preserved planning performance.

\begin{table}[!t]
    \centering
    \caption{\textbf{Ablation study of latency in \texttt{NAVSIMv1} \cite{dauner2024navsim}}. `World Model' includes action-conditioned future-state prediction and future-aware utility readout. `Action Head' includes native utility scoring, paired candidate comparison and score refinement.
    }
    \label{tab:ablation_module_latency}
    \setlength{\tabcolsep}{7pt}
    \renewcommand{\arraystretch}{0.5}
    \resizebox{\columnwidth}{!}{%
    \begin{tabular}{l|ccccc}
        \toprule
        \multirow{2}{*}{Module}&Visual&Prop. Update&Action&World&\multirow{2}{*}{Total}\\
        &Encoding&(4$\times$)&Head&Model&\\
        \midrule
        \midrule
        Latency (ms) & 5.63 & $\sim$0.40 & 0.13 & 0.05 &\cellcolor{green!20} \textbf{7.55} \\
        \bottomrule
    \end{tabular}%
    }
\end{table}

\begin{table}[!t]
    \centering
    \caption{\textbf{Ablation study of
     the number of proposals in \texttt{NAVSIMv1} \cite{dauner2024navsim}.} Default setting of FastWorld in green.}
    \label{tab:ablation_candidate_budget}
    \setlength{\tabcolsep}{3pt}
    \renewcommand{\arraystretch}{0.5}
    \resizebox{\columnwidth}{!}{%
    \begin{tabular}{cc|cccccc|c}
        \toprule
        Total Prop. & Refined Prop.& NC$\uparrow$ & DAC$\uparrow$ & TTC$\uparrow$ & C$\uparrow$ & EP$\uparrow$ & PDMS$\uparrow$ & FPS$\uparrow$ \\
        \midrule
        \midrule
        32 & 16 & 98.4 & 98.2 & 94.5 & 99.9 & 89.4 & 92.1 & 134.2 \\
        \cellcolor{green!20}\textbf{64} & \cellcolor{green!20}\textbf{16} & 98.5 & 98.0 & 94.6 & 99.9 & 89.9 & 92.2 & 132.4 \\
        64 & 32 & 98.5 & 98.0 & 94.6 & 99.9 & 89.8 & 92.2 & 129.8 \\
        128 & 16 & 95.7 & 96.5 & 87.7 & 97.3 & 88.4 & 87.0 & 123.7 \\
        128 & 32 & 95.7 & 96.7 & 87.8 & 97.2 & 88.5 & 87.2 & 123.2 \\
        128 & 64 & 95.6 & 96.7 & 88.0 & 97.2 & 88.4 & 87.2 & 122.4 \\
        \bottomrule
    \end{tabular}%
    }
\end{table}

\begin{table}[!t]
    \centering
    \caption{\textbf{Ablation study of world state prediction horizon in \texttt{NAVSIMv1}~\cite{dauner2024navsim}.} Default setting of FastWorld in green. `None' indicates that the world state prediction is replaced by a random tensor of the same shape.}
    \label{tab:ablation_prediction_horizon}
    \setlength{\tabcolsep}{6pt}
    \renewcommand{\arraystretch}{0.5}
    \resizebox{\columnwidth}{!}{%
    \begin{tabular}{c|cccccc|c}
        \toprule
        Pred. Horizon & NC$\uparrow$ & DAC$\uparrow$ & TTC$\uparrow$ & C$\uparrow$ & EP$\uparrow$ & PDMS$\uparrow$ & FPS$\uparrow$ \\
        \midrule
        \midrule
        None  & 98.2 & 97.7 & 94.2 & 99.8 & 89.9 & 91.8 & 133.3 \\
        \cellcolor{green!20}\textbf{1s}  & 98.5 & 98.0 & 94.6 & 99.9 & 89.9 & 92.2 & 132.4 \\
        2s & 98.5&	98.0	&94.7&	99.9	&89.8	&92.2	 & 130.1 \\
        3s & 98.3&	97.8&	94.2&	99.9&	90.1&	92.0& 129.3 \\
        4s & 98.6&	98.0&	94.8&	99.9&	89.6&	92.2 & 128.6 \\
        \bottomrule
    \end{tabular}%
    }
\end{table}

\begin{table}[!t]
    \centering
    \caption{\textbf{Ablation study of visual encoder in \texttt{NAVSIMv1}~\cite{dauner2024navsim}.} Default setting of FastWorld in green.}
    \label{tab:ablation_backbone}
    \setlength{\tabcolsep}{3pt}
    \renewcommand{\arraystretch}{0.5}
    \resizebox{\columnwidth}{!}{%
    \begin{tabular}{c|c|cccccc|c}
        \toprule
        Name & Param.$\downarrow$& NC$\uparrow$ & DAC$\uparrow$ & TTC$\uparrow$ & C$\uparrow$ & EP$\uparrow$ & PDMS$\uparrow$ & FPS$\uparrow$ \\
        \midrule
        \midrule
        \cellcolor{green!20}\textbf{ResNet-34} & 27.0 & 98.5 & 98.0 & 94.6 & 99.9 & 89.9 & 92.2 & 132.4 \\
        ViT-L&309.2 & 97.0 & 95.6 & 90.3 & 99.9 & 88.6 & 88.3 & 20.5 \\
        \bottomrule
    \end{tabular}%
    }
\end{table}

\subsection{Ablation Study}

\noindent \textbf{Ablation of latency.}
Tab.~\ref{tab:ablation_module_latency} presents the latency of FastWorld.
The world model takes only 0.05\,ms, contributing minimally to the total inference latency.

\noindent \textbf{Ablation of the number of proposals.}
Tab.~\ref{tab:ablation_candidate_budget} presents the results with different numbers of total and refined proposals.
The first three configurations all achieve satisfactory performance, with 64 total proposals and 16 refined proposals providing the best balance between efficiency and performance.
Increasing the proposal count to 128 impairs trajectory selection, reducing both PDMS and FPS.

\noindent \textbf{Ablation of the world state prediction horizon.}
Tab.~\ref{tab:ablation_prediction_horizon} presents the results with different world state prediction horizons.
Longer horizons do not consistently improve performance but steadily reduce FPS, with the world model latency increasing from 0.05ms (see Tab.~\ref{tab:ablation_module_latency}) at 1s to 0.18ms at 4s.
The FPS degradation is modest, as the world model contributes little to total inference latency.

\noindent \textbf{Ablation of visual encoder.}
Tab.~\ref{tab:ablation_backbone} presents the results with different visual encoders.
Replacing ResNet-34 with ViT-L increases the parameter count but reduces PDMS from 92.2 to 88.3 and FPS from 132.4 to 20.5.
These results suggest that, unlike Drive-JEPA~\cite{wang2026drivejepa}, our lightweight design does not fully exploit the potential of such an expensive visual representation.

See the Supplementary Material for additional ablation studies.

\subsection{Visualization}

\begin{figure}[!t]
\centering
\includegraphics[width=1.\linewidth]{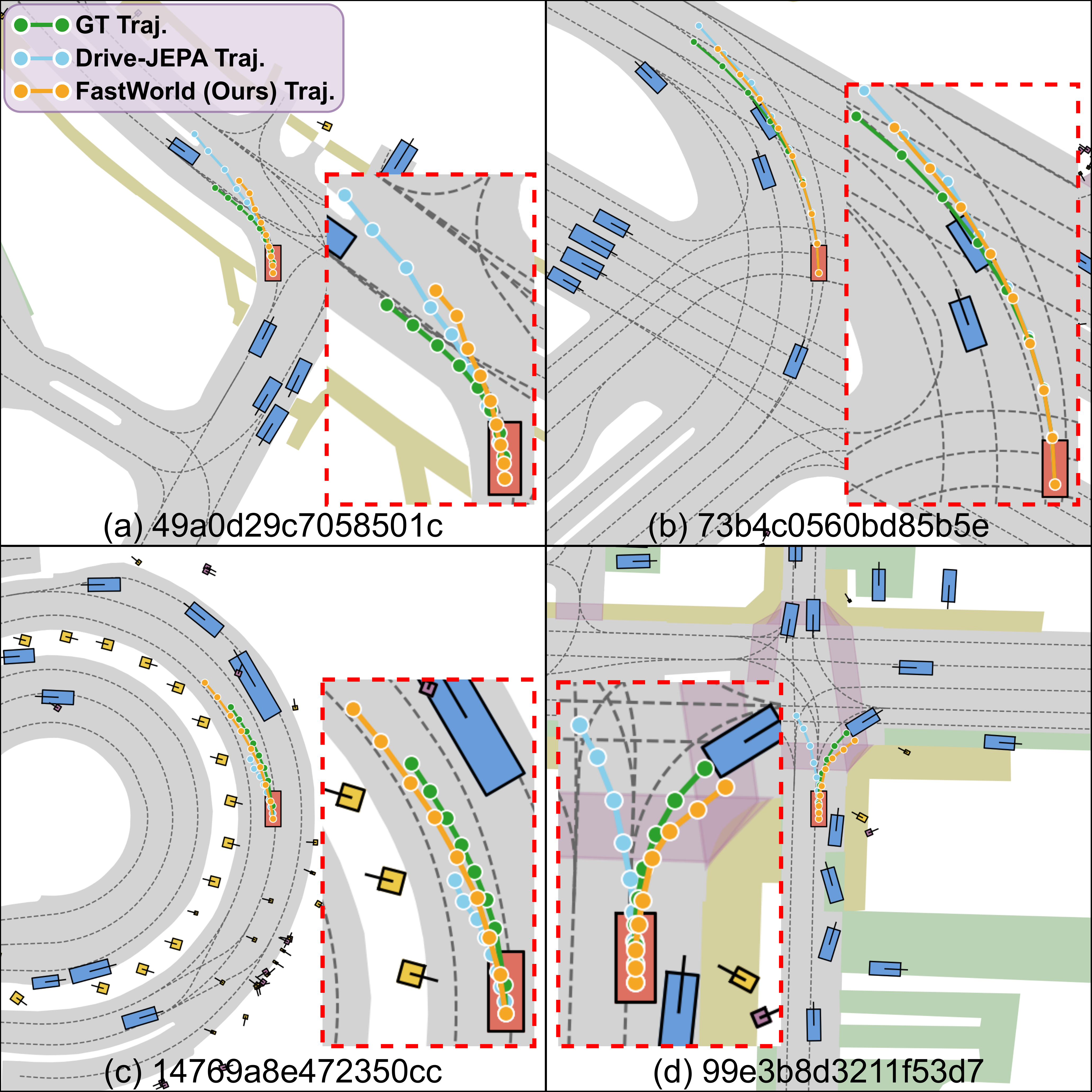}
\caption{\textbf{4s trajectory planning visualization examples of turning scenarios from \texttt{NAVSIM}~\cite{dauner2024navsim} for FastWorld and Drive-JEPA~\cite{wang2026drivejepa}}. Subfigure label: sample tokens. Top label: source of trajectory.}
\label{fig:vis}
\end{figure}

As shown in Fig.~\ref{fig:vis}, we provide some visualization examples of turning scenarios from \textbf{\texttt{NAVSIM}}~\cite{dauner2024navsim}. FastWorld achieves stable turning trajectories with better alignment to the GT compared with Drive-JEPA~\cite{wang2026drivejepa}. See the Supplementary Material for more visualization. 

\section{Conclusion}
\label{sec:conclusion}
In this study, we introduce \textbf{\textit{FastWorld}}, an efficient autonomous driving framework that integrates compact world prediction into trajectory selection. Specifically, Action-Conditioned World Prediction forecasts compact future representations for candidate actions in parallel, while Future-Aware Utility Scoring uses these predictions to refine candidate rankings. By directly predicting future states in a compact latent space, FastWorld incorporates future scene information into decision-making with minimal computational overhead. Evaluations on NAVSIM and nuScenes demonstrate that FastWorld achieves leading inference speed while maintaining competitive planning performance, offering practical insights into the efficient deployment of world models for autonomous driving.


\section*{LLM Usage Statement}
ChatGPT was used to assist with code drafting, and Doubao was employed to aid manuscript preparation, including translation, polishing and grammatical refinement. All method design, literature organization, figures, mathematical formulations, experimental analysis, and all technical claims were independently developed, written, and verified by the authors. No generative AI tools produced core technical contributions, experimental data, or evaluation results.

\section*{Acknowledgements}
This work is supported by the Fundamental Research Funds for the Central Universities (No. 2024XKRC024, Key Program), the National Natural Science Foundation of China (NSFC) under Grants No. 62536001 and No. 62576026, and the Beijing Nova Program (202604841209).
We would like to thank Caixia Wu for insightful discussions and helpful feedback throughout the writing process.

{
    \small
    \bibliographystyle{ieeenat_fullname}
    \bibliography{main}
}

\clearpage

\maketitlesupplementary

\section*{S1. More Details about Training Objective}
\label{sec:supp_loss}

The training objective combines future-state prediction, candidate ranking, expected utility maximization, and planning regularization:
\begin{equation}
\mathcal{L}
=
1\cdot\mathcal{L}_{\mathrm{obs}}
+0.1\cdot\mathcal{L}_{\mathrm{rank}}
+0.2\cdot\mathcal{L}_{\mathrm{util}}
+\mathcal{L}_{\mathrm{reg}}
\tag{S1}
\end{equation}

Here, we detail the regularization term $\mathcal{L}_{\mathrm{reg}}$, which contains residual,
imitation, and planner consistency losses:
\begin{equation}
\begin{aligned}
\mathcal{L}_{\mathrm{reg}}
={}&0.002\cdot\mathcal{L}_{\mathrm{res}}
+0.01\cdot\mathcal{L}_{\mathrm{imit}}\\
&+200\cdot\mathcal{L}_{\mathrm{compat}}
+20\cdot\mathcal{L}_{\mathrm{native}}
+20\cdot\mathcal{L}_{\mathrm{prior}}
\end{aligned}
\tag{S2}
\end{equation}

\noindent\textbf{Residual regularization.}
The future-aware correction for candidate $i$ is $f_{\mathrm{util}}(\hat z_{t+\Delta}^{\,i})$, which contributes to its final score $s_i$ together with the comparison prior $b_i$.
Let ${\bar{\mathcal{F}}}_{\mathrm{util}}$ denote the mean correction over the retained
candidate set $\mathcal{S}$. We penalize the centered future-aware corrections as $\mathcal{L}_{\mathrm{res}}$:
\begin{equation}
\mathcal{L}_{\mathrm{res}}
=
\frac{1}{K}\sum_{i\in\mathcal{S}}
\left(
f_{\mathrm{util}}(\hat z_{t+\Delta}^{\,i})
-{\bar{\mathcal{F}}}_{\mathrm{util}}
\right)^2
\tag{S3}
\end{equation}
where $K$ is the number of retained candidates. Centering removes the mean correction, so this term constrains relative score changes while leaving a shared offset unpenalized.

\noindent\textbf{Trajectory imitation.}
For each proposal $\tau_i \in \mathcal{T}^{(R)}$, with pose $\tau_{i,h}$ at planning step $h$, let $\operatorname{pos}(\tau_{i,h})$ denote its position components, and let $\mathbf{x}^{\mathrm{log}}_h$ denote the logged position at that step. The best-of-$N$ imitation loss $\mathcal{L}_{\mathrm{imit}}$ is:
\begin{equation}
\mathcal{L}_{\mathrm{imit}}
=
\min_{i\in\{1,\ldots,N\}}
\frac{1}{H}\sum_{h=1}^{H}
\mathcal{L}_{\mathrm{sm}}
\bigl(\operatorname{pos}(\tau_{i,h}),
\mathbf{x}^{\mathrm{log}}_h\bigr)
\tag{S4}
\end{equation}
where $N$ is the total number of proposals, $H$ is the number of planning steps, and $\mathcal{L}_{\mathrm{sm}}(\cdot)$ denotes the Smooth L1 loss. The minimum selects the proposal with the
smallest imitation loss, encouraging the proposal set to cover the
logged trajectory.

\noindent\textbf{Proposal compatibility.}
For each proposal $\tau_i\in\mathcal{T}^{(R)}$, before the router selects the retained indices $\mathcal{S}$, the initialized planner provides a query-matched proposal $\tau_i^{0}$ and basic readout score $\ell_i^{0}$. We denote its full proposal set by $\mathcal{T}^{(R),0}=\{\tau_i^{0}\}_{i=1}^{N}$. The current and initialized coarse utility probabilities are $q_i=\operatorname{sigmoid}(\ell_i)$ and $q_i^{0}=\operatorname{sigmoid}(\ell_i^{0})$, respectively. The proposal compatibility loss $\mathcal{L}_{\mathrm{compat}}$ is:
\begin{equation}
\begin{aligned}
\mathcal{L}_{\mathrm{compat}}
={}&
\mathcal{L}_{\mathrm{sm}}
\bigl(\mathcal{T}^{(R)},\mathcal{T}^{(R),0}\bigr)\\
&+
\frac{1}{N}\sum_{i=1}^{N}(q_i-q_i^{0})^2
\end{aligned}
\tag{S5}
\end{equation}
where $N$ is the total number of proposals, and $\mathcal{L}_{\mathrm{sm}}(\cdot)$ denotes the Smooth L1 loss. This loss keeps the full proposal set and its coarse utility probabilities close to those of the initialized planner, preserving its proposal and scoring prior during joint training.

\noindent\textbf{Native-score consistency.}
For each proposal $\tau_i\in\mathcal{T}^{(R)}$, the native-score consistency loss $\mathcal{L}_{\mathrm{native}}$ is:
\begin{equation}
\mathcal{L}_{\mathrm{native}}
=
\frac{1}{N}\sum_{i=1}^{N}
(\ell_i-\ell_i^{0})^2
\tag{S6}
\end{equation}
where $N$ is the total number of proposals. This loss anchors the basic readout scores of all proposals to their initialized values, stabilizing the base scoring branch during joint training.

\noindent\textbf{Comparison-prior consistency.}
After candidate selection, the comparison prior $b_i$ is computed for
each retained candidate $i\in\mathcal{S}$. The initialized planner has
its own retained index set $\mathcal{S}^{0}$ and corresponding prior
scores $b_i^{0}$. We compare the priors only for query indices retained
by both planners:
\begin{equation}
\mathcal{L}_{\mathrm{prior}}
=
\frac{
\sum_{i\in\mathcal{S}\cap\mathcal{S}^{0}}
(b_i-b_i^{0})^2
}{
\max\bigl(1,|\mathcal{S}\cap\mathcal{S}^{0}|\bigr)
}
\tag{S7}
\end{equation}
Here, $\mathcal{S}^{0}$ is the initialized planner's retained index set, and $b_i^{0}$ is its comparison prior for query index $i$.
Candidates are matched by their original query indices. The loss is 0 when the intersection is empty. These consistency losses compare the current planner with its own initialization and require no external teacher.

\section*{S2. More Details about Dataset}
\label{sec:supp_dataset}

\noindent\textbf{\texttt{NAVSIMv1}.}
NAVSIMv1~\cite{dauner2024navsim} is built on OpenScene~\cite{openscene2023_official}, a 2Hz redistribution of nuPlan~\cite{caesar2021nuplan}, which contains approximately 120 hours of urban driving data. Each sample provides eight camera views and a fused LiDAR point cloud from five sensors. The benchmark provides filtered \texttt{navtrain} and \texttt{navtest} splits and evaluates a predicted four-second trajectory in a non-reactive simulation: the ego vehicle follows the plan using an LQR controller and a kinematic bicycle model, while other actors replay their recorded future motion. The primary metric Predictive Driver Model Score (PDMS) is:
\begin{equation}
\mathrm{PDMS}
=
\mathrm{NC}\cdot\mathrm{DAC}
\cdot
\frac{5\,\mathrm{EP}+5\,\mathrm{TTC}+2\,\mathrm{C}}{12}
\tag{S8}
\end{equation}
where NC is No At-Fault Collision, DAC is Drivable Area Compliance, EP is Ego Progress, TTC is Time to Collision, and C is Comfort. NC and DAC act as multiplicative safety penalties. EP measures progress along the route relative to a safe-progress reference, TTC penalizes a predicted time to collision below the safety threshold, and C evaluates trajectory comfort from kinematic quantities such as acceleration and jerk. All component scores are normalized to $[0,1]$. PDMS is averaged over evaluated frames and reported on a $0$--$100$ scale.

\noindent\textbf{\texttt{NAVSIMv2}.}
NAVSIMv2~\cite{cao2025pseudo_navsimv2} extends the OpenScene-based benchmark with pseudo-simulation, including pre-generated synthetic follow-up observations for two-stage evaluation. The primary metric Extended Predictive Driver Model Score (EPDMS), extends PDMS with additional rule-compliance and comfort terms:
\begin{equation}
\begin{aligned}
\mathrm{EPDMS}
=&
\tilde{s}_{\mathrm{NC}}
\cdot\tilde{s}_{\mathrm{DAC}}
\cdot\tilde{s}_{\mathrm{DDC}}
\cdot\tilde{s}_{\mathrm{TLC}}
\\
&\cdot\frac{
5\tilde{s}_{\mathrm{TTC}}
+5\tilde{s}_{\mathrm{EP}}
+2\tilde{s}_{\mathrm{LK}}
+2\tilde{s}_{\mathrm{HC}}
+2\tilde{s}_{\mathrm{EC}}
}{16}
\end{aligned}
\tag{S9}
\end{equation}
where NC, DAC, EP, and TTC retain their NAVSIMv1~\cite{dauner2024navsim} meanings; DDC is Driving Direction Compliance, TLC is Traffic Light Compliance, LK is Lane Keeping, HC is History Comfort, and EC is Extended Comfort. The multiplicative terms penalize collisions, leaving the drivable area, wrong-way driving, and red-light violations. The weighted terms measure progress, collision risk, lane keeping, comfort across the transition from human history to the predicted plan, and consistency between successive plans. Each $\tilde{s}_m$ is the filtered score for metric $m$: it equals $1$ when the human trajectory also violates that metric, and otherwise equals the agent's score. EPDMS is computed from normalized scores in $[0,1]$ and reported on a $0$--$100$ scale. The \textbf{\texttt{navtest}} split uses the single-stage EPDMS evaluation. For the \textbf{\texttt{navhard}} two-stage split, each initial scene and its synthetic follow-up scenes are scored with EPDMS. Follow-up scores are weighted by their spatial proximity to the endpoint predicted in the first stage, and the weighted second-stage score is multiplied by the first-stage score:
\begin{equation}
\mathrm{EPDMS}_{\mathrm{2-stage}}
=
\mathrm{EPDMS}_{\mathrm{s_{1}}} \sum_j w_j\,\mathrm{EPDMS}_{\mathrm{s_{2}}}^{j}
\tag{S10}
\end{equation}
where $\mathrm{EPDMS}_{\mathrm{s_{1}}}$ is the initial-scene score, $\mathrm{EPDMS}_{\mathrm{s_{2}}}^{j}$ is the score for follow-up scene $j$, and $w_j$ is its normalized proximity weight. 

\noindent\textbf{\texttt{nuScenes}.}
nuScenes~\cite{NUSCENES} contains 1,000 annotated scenes collected in Boston and Singapore. Each scene spans 20 seconds and is annotated at 2Hz, with data from six cameras, one LiDAR, and five radars. The first metric is the L2, computed at prediction horizons $t\in\{1,2,3\}$ seconds:
\begin{equation}
\mathrm{L2}(t)
=
\frac{1}{N}\sum_{n=1}^{N}
\left\|
\hat{\mathbf{p}}_n(t)
-\mathbf{p}^{\mathrm{gt}}_n(t)
\right\|_2
\tag{S11}
\end{equation}
where $N$ is the number of evaluated samples, and $\hat{\mathbf{p}}_n(t)$ and $\mathbf{p}^{\mathrm{gt}}_n(t)$ are the predicted and logged ego positions for sample $n$ at time $t$. L2 is reported in meters, and lower values indicate more accurate trajectory prediction. The second metric is Collision Rate (CR), which measures the fraction of evaluated samples whose predicted ego trajectory collides with another actor by horizon $t$:
\begin{equation}
\mathrm{CR}(t)
=
\frac{100}{N H_t}
\sum_{n=1}^{N}\sum_{h=1}^{H_t} c_{n,h}.
\tag{S12}
\end{equation}
where $N$ is the number of valid evaluation samples, $H_t$ is the
number of waypoints up to horizon $t$, and $c_{n,h}$ is the box
collision indicator for sample $n$ at waypoint $h$.

\section*{S3. More Details about Implementation}
\label{sec:supp_implementation}

\noindent\textbf{\texttt{NAVSIM}.}
The ResNet-34~\cite{he2016deep_resnet} FPN outputs a $256\times14\times24$ feature map per camera at stride 32. We flatten the spatial grid into 336 feature vectors per view and preserve the four views for image cross-attention. For NAVSIMv1, we jointly fine-tune the planner and the world model for 30 epochs on 8 NVIDIA GeForce RTX 4090 GPUs with a global batch size of 128. The approximate wall-clock training time is 75 hours. The planner uses a lower learning rate than the future prediction module.

\noindent\textbf{\texttt{nuScenes}.}
We project the 768-channel Swin-T~\cite{liu2021swin_transformer} features to 256 channels and use six learned queries per camera view to obtain a $36\times256$ scene representation for each frame. We jointly train the model on images for 15 epochs, then continue training for 80 epochs using cached scene representations and trajectory proposals, with an overall wall-clock training time of approximately 20 hours. Retrieving the cached features by timestamp avoids repeated image encoding and proposal generation. During the final three epochs, we keep the world dynamics fixed and update the trajectory scorer.

\section*{S4. More Ablation Study}
\label{sec:supp_ablation}

\noindent \textbf{Ablation of proposal pairing.}
Tab.~\ref{tab:supp_ablation_navsim_pairing} compares two natural ways to pair proposals. Score rank pairs candidates with different utility scores, while trajectory divergence pairs candidates with different paths. The two rules differ by only 0.1 PDMS and both run at around 130 FPS.

\noindent \textbf{Ablation of latency in \texttt{nuScenes}~\cite{NUSCENES}.}
Tab.~\ref{tab:supp_ablation_nus_latency} reports the inference latency of FastWorld on \textbf{\texttt{nuScenes}}. Compared with the \textbf{\texttt{NAVSIM}} results in Tab.~5 of the main paper, most of the additional latency comes from visual encoding and the world model. The visual backbone encodes both historical and current surround-view images. The world model takes 8.54 ms as it predicts a waypoint-conditioned state for each input frame and applies a Transformer transition over three rollout steps for the 16 shortlisted proposals. In comparison, the \textbf{\texttt{NAVSIM}} version uses a single lightweight transition to predict the target-time state directly.

\noindent \textbf{Ablation of the number of proposals in \texttt{nuScenes}~\cite{NUSCENES}.} Tab.~\ref{tab:supp_ablation_nus_proposals} compares different numbers of refined proposals on \texttt{nuScenes}, with the total fixed at 64. Refining 8 or 16 proposals yields similar L2, collision rates, and inference speeds. A separate run with 32 refined proposals also yields nearly identical results to the default setting. Since the proposals are generated around a shared waypoint sequence, a small shortlist appears sufficient for this setting.

\noindent \textbf{Ablation of displacement scoring in \texttt{nuScenes}~\cite{NUSCENES}.}
Tab.~\ref{tab:supp_ablation_nus_displacement} evaluates displacement scoring, an auxiliary scoring term specific to the \textbf{\texttt{nuScenes}} variant of FastWorld. On \textbf{\texttt{NAVSIM}}, the PDMS supervision already includes EP. On \textbf{\texttt{nuScenes}}, proposals can follow similar paths yet incur different L2 if they cover different distances over the same time. To account for this difference during ranking, displacement scoring compares each proposal's displacement magnitude at 1, 2, and 3 seconds with the corresponding predicted distribution of ego displacement. Adding this term reduces the average L2 from 0.47m to 0.40m and the average collision rate from 0.34\% to 0.31\%, while inference speed decreases slightly from 16.9 to 16.6 FPS.

\noindent \textbf{Ablation of world state prediction in \texttt{nuScenes}~\cite{NUSCENES}.}
Tab.~\ref{tab:supp_ablation_nus_future_state} shows that predicted world state reduce L2 at all three horizons compared with random world state. The average L2 drops from 0.51m to 0.40m, while inference speed decreases from 18.0 to 16.6 FPS.

\begin{table}[!t]
    \centering
    \renewcommand{\thetable}{S1}
    \caption{\textbf{Ablation study of proposal pairing in \texttt{NAVSIMv1}~\cite{dauner2024navsim}.} `Score Rank' pairs candidates based on their scores, while `Trajectory Divergence' pairs candidates based on differences in their paths. Default setting of FastWorld in green.}
    \label{tab:supp_ablation_navsim_pairing}
    \setlength{\tabcolsep}{3pt}
    \renewcommand{\arraystretch}{0.5}
    \resizebox{\columnwidth}{!}{%
    \begin{tabular}{c|cccccc|c}
        \toprule
        Pairing & NC$\uparrow$ & DAC$\uparrow$ & TTC$\uparrow$ & C$\uparrow$ & EP$\uparrow$ & PDMS$\uparrow$ & FPS$\uparrow$ \\
        \midrule
        \midrule
        \cellcolor{green!20}\textbf{Score Rank} & 98.5 & 98.0 & 94.6 & 99.9 & 89.9 & 92.2 & 132.4 \\
        Trajectory Divergence & 98.4 & 97.9 & 94.6 & 99.9 & 89.8 & 92.1 & 129.7 \\
        \bottomrule
    \end{tabular}%
    }
\end{table}

\begin{table}[!t]
    \centering
    \renewcommand{\thetable}{S2}
    \caption{\textbf{Ablation study of latency in \texttt{nuScenes} \cite{NUSCENES}}. `World Model' includes action-conditioned future-state prediction and future-aware utility readout. `Prop. Update' includes one waypoint decoding pass per frame and the generation of 64 proposals from the current-frame waypoints. `Action Head' includes native utility scoring, paired candidate comparison and score refinement.
    }
    \label{tab:supp_ablation_nus_latency}
    \setlength{\tabcolsep}{7pt}
    \renewcommand{\arraystretch}{0.5}
    \resizebox{\columnwidth}{!}{%
    \begin{tabular}{l|ccccc}
        \toprule
        \multirow{2}{*}{Module}&Visual&\multirow{2}{*}{Prop. Update}&Action&World&\multirow{2}{*}{Total}\\
        &Encoding&&Head&Model&\\
        \midrule
        \midrule
        Latency (ms) & 45.72 & 1.65 & 3.28 & 8.54 & \cellcolor{green!20} \textbf{60.09} \\
        \bottomrule
    \end{tabular}%
    }
\end{table}

\begin{table}[!t]
    \centering
    \renewcommand{\thetable}{S3}
    \caption{\textbf{Ablation study of
     the number of proposals in \texttt{nuScenes} \cite{NUSCENES}.} Default setting of FastWorld in green.}
    \label{tab:supp_ablation_nus_proposals}
    \setlength{\tabcolsep}{2pt}
    \renewcommand{\arraystretch}{0.5}
    \resizebox{\columnwidth}{!}{%
    \begin{tabular}{cc|cccc|cccc|c}
    \toprule
    \multirow{2.5}{*}{Total Prop.}&\multirow{2.5}{*}{Refined Prop.}&\multicolumn{4}{c|}{L2(m)$\downarrow$}&\multicolumn{4}{c|}{CR(\%)$\downarrow$}&\multirow{2.5}{*}{FPS$\uparrow$} \\
    \cmidrule{3-10}
    &&1s&2s&3s&Avg.&1s&2s&3s&Avg.&\\
    \midrule
    \midrule
    64&8 & 0.22 & 0.38 & 0.63 & 0.41 & 0.12 & 0.32 & 0.54 & 0.33 & 16.6\\
    \cellcolor{green!20} \textbf{64} &\cellcolor{green!20} \textbf{16} & 0.21 & 0.37 & 0.63 & 0.40 & 0.10 & 0.30 & 0.52 & 0.31 & 16.6 \\
    \bottomrule
    \end{tabular}%
    }
\end{table}

\begin{table}[!t]
    \centering
    \renewcommand{\thetable}{S4}
    \caption{\textbf{Ablation study of displacement scoring in \texttt{nuScenes}~\cite{NUSCENES}.} Default setting of FastWorld in green.}
    \label{tab:supp_ablation_nus_displacement}
    \setlength{\tabcolsep}{3pt}
    \renewcommand{\arraystretch}{0.5}
    \resizebox{\columnwidth}{!}{%
    \begin{tabular}{c|cccc|cccc|c}
    \toprule
    \multirow{2.5}{*}{Displacement}&\multicolumn{4}{c|}{L2(m)$\downarrow$}&\multicolumn{4}{c|}{CR(\%)$\downarrow$}&\multirow{2.5}{*}{FPS$\uparrow$} \\
    \cmidrule{2-9}
    &1s&2s&3s&Avg.&1s&2s&3s&Avg.&\\
    \midrule
    \midrule
    \cellcolor{green!20} \textbf{w/ Displacement} & 0.21 & 0.37 & 0.63 & 0.40 & 0.10 & 0.30 & 0.52 & 0.31 & 16.6 \\
    w/o Displacement & 0.25 & 0.44 & 0.71 & 0.47 & 0.10 & 0.32 & 0.60 & 0.34 & 16.9\\
    \bottomrule
    \end{tabular}%
    }
\end{table}

\begin{table}[!t]
    \centering
    \renewcommand{\thetable}{S5}
    \caption{\textbf{Ablation study of world state prediction in \texttt{nuScenes}~\cite{NUSCENES}.} `Random' replaces the predicted states with a fixed, same-shape Gaussian tensor before utility readout. Default setting of FastWorld in green.}
    \label{tab:supp_ablation_nus_future_state}
    \setlength{\tabcolsep}{4pt}
    \renewcommand{\arraystretch}{0.5}
    \resizebox{\columnwidth}{!}{%
    \begin{tabular}{c|cccc|cccc|c}
    \toprule
    \multirow{2.5}{*}{Future State}&\multicolumn{4}{c|}{L2(m)$\downarrow$}&\multicolumn{4}{c|}{CR(\%)$\downarrow$}&\multirow{2.5}{*}{FPS$\uparrow$} \\
    \cmidrule{2-9}
    &1s&2s&3s&Avg.&1s&2s&3s&Avg.&\\
    \midrule
    \midrule
    \cellcolor{green!20} \textbf{Predicted} & 0.21 & 0.37 & 0.63 & 0.40 & 0.10 & 0.30 & 0.52 & 0.31 & 16.6 \\
    Random & 0.26 & 0.48 & 0.79 & 0.51 & 0.15 & 0.25 & 0.50 & 0.30 & 18.0\\
    \bottomrule
    \end{tabular}%
    }
\end{table}

\section*{S5. More Visualization}
\label{sec:supp_vis}

\begin{figure*}[!t]
\renewcommand\thefigure{S1}
\centering
\includegraphics[width=0.92\linewidth]{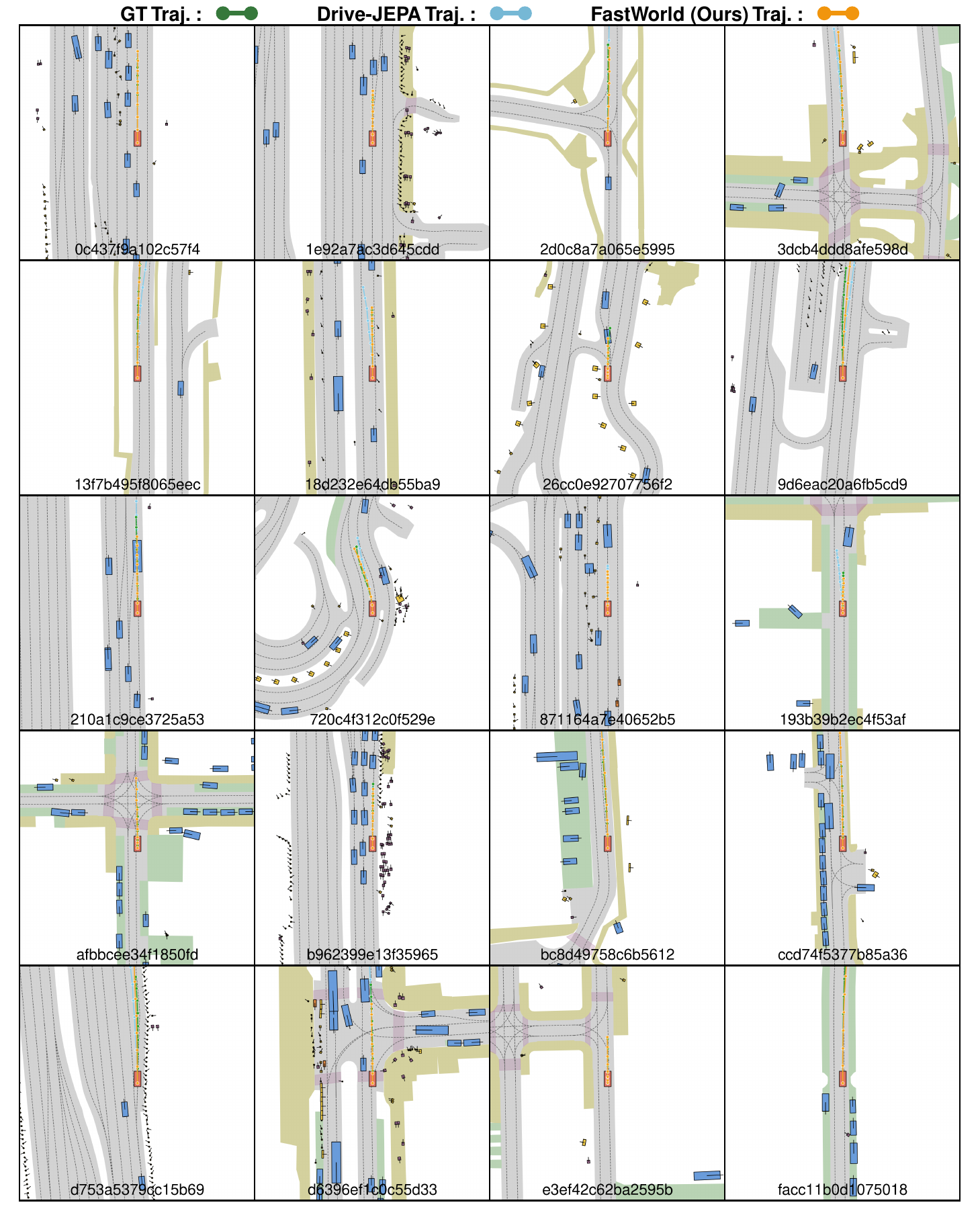}
\caption{\textbf{Visualization examples of \texttt{NAVSIM}~\cite{dauner2024navsim} on straight trajectory.} Subfigure label: sample tokens. Top label: source of trajectory.}
\label{fig:supp_navsim_vis1}
\end{figure*}

\begin{figure*}[!t]
\renewcommand\thefigure{S2}
\centering
\includegraphics[width=0.92\linewidth]{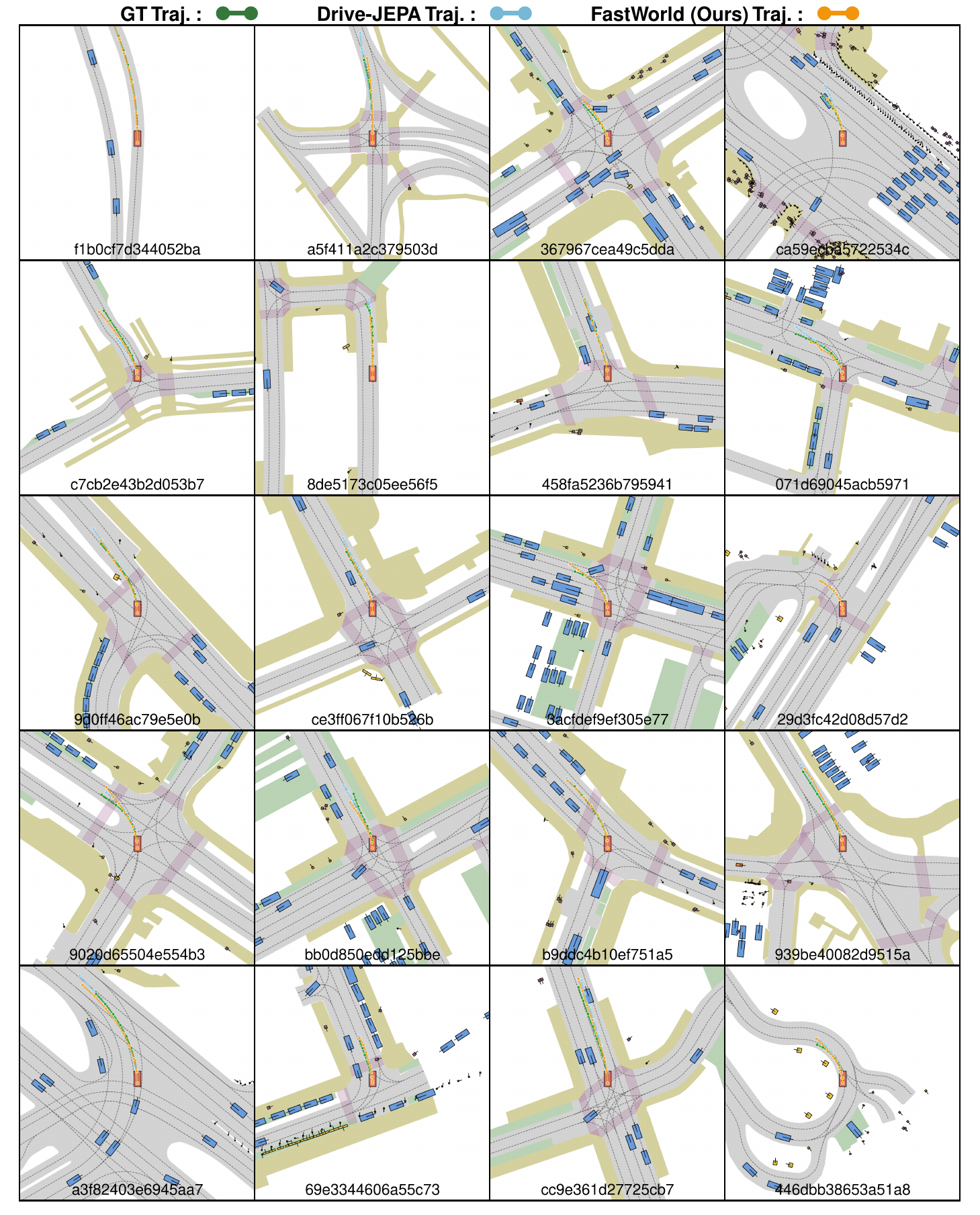}
\caption{\textbf{Visualization examples of \texttt{NAVSIM}~\cite{dauner2024navsim} on left-turn trajectory.} Subfigure label: sample tokens. Top label: source of trajectory.}
\label{fig:supp_navsim_vis2}
\end{figure*}

\begin{figure*}[!t]
\renewcommand\thefigure{S3}
\centering
\includegraphics[width=0.92\linewidth]{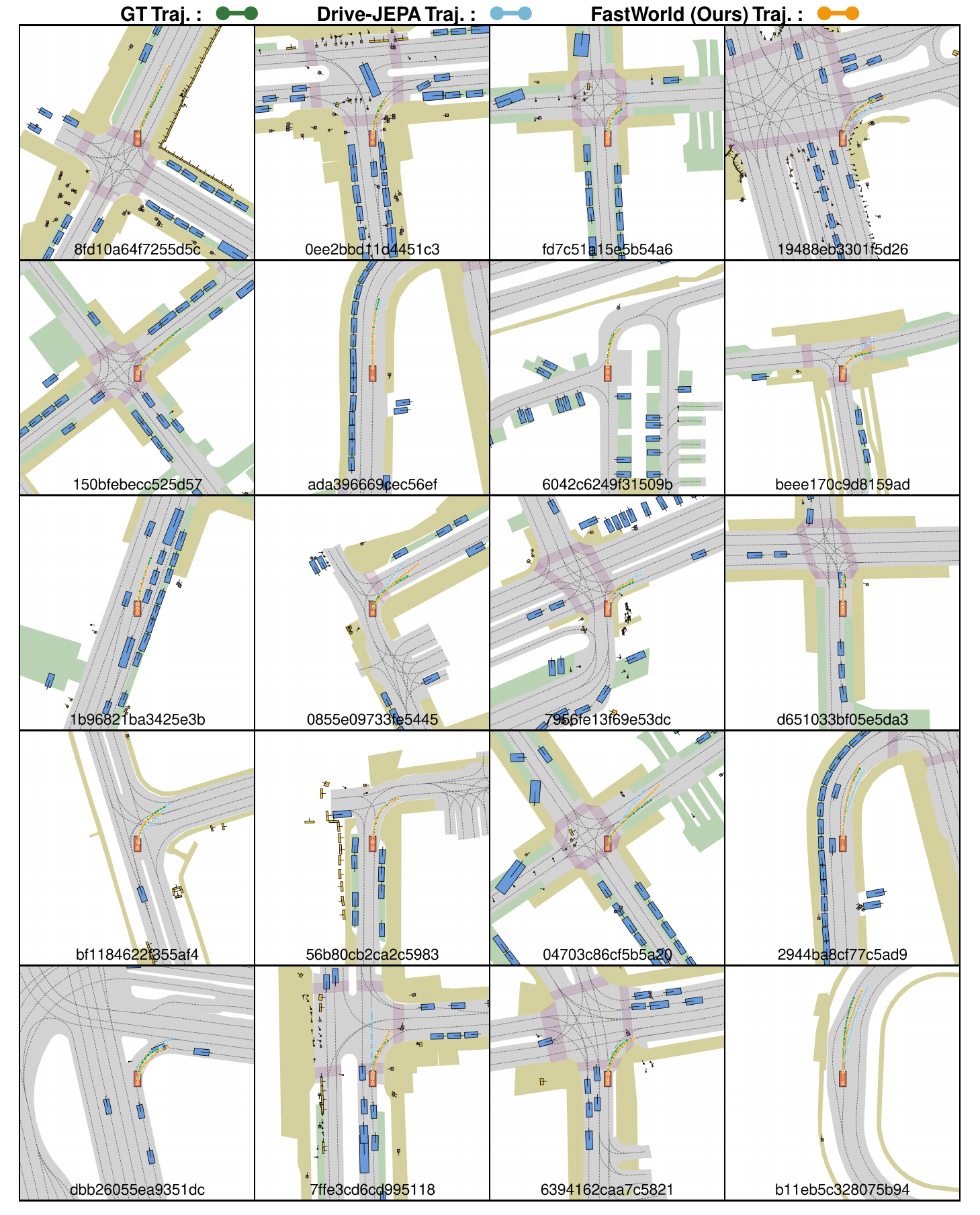}
\caption{\textbf{Visualization examples of \texttt{NAVSIM}~\cite{dauner2024navsim} on right-turn trajectory.} Subfigure label: sample tokens. Top label: source of trajectory.}
\label{fig:supp_navsim_vis3}
\end{figure*}

\begin{figure*}[!t]
\renewcommand\thefigure{S4}
\centering
\includegraphics[width=1\linewidth]{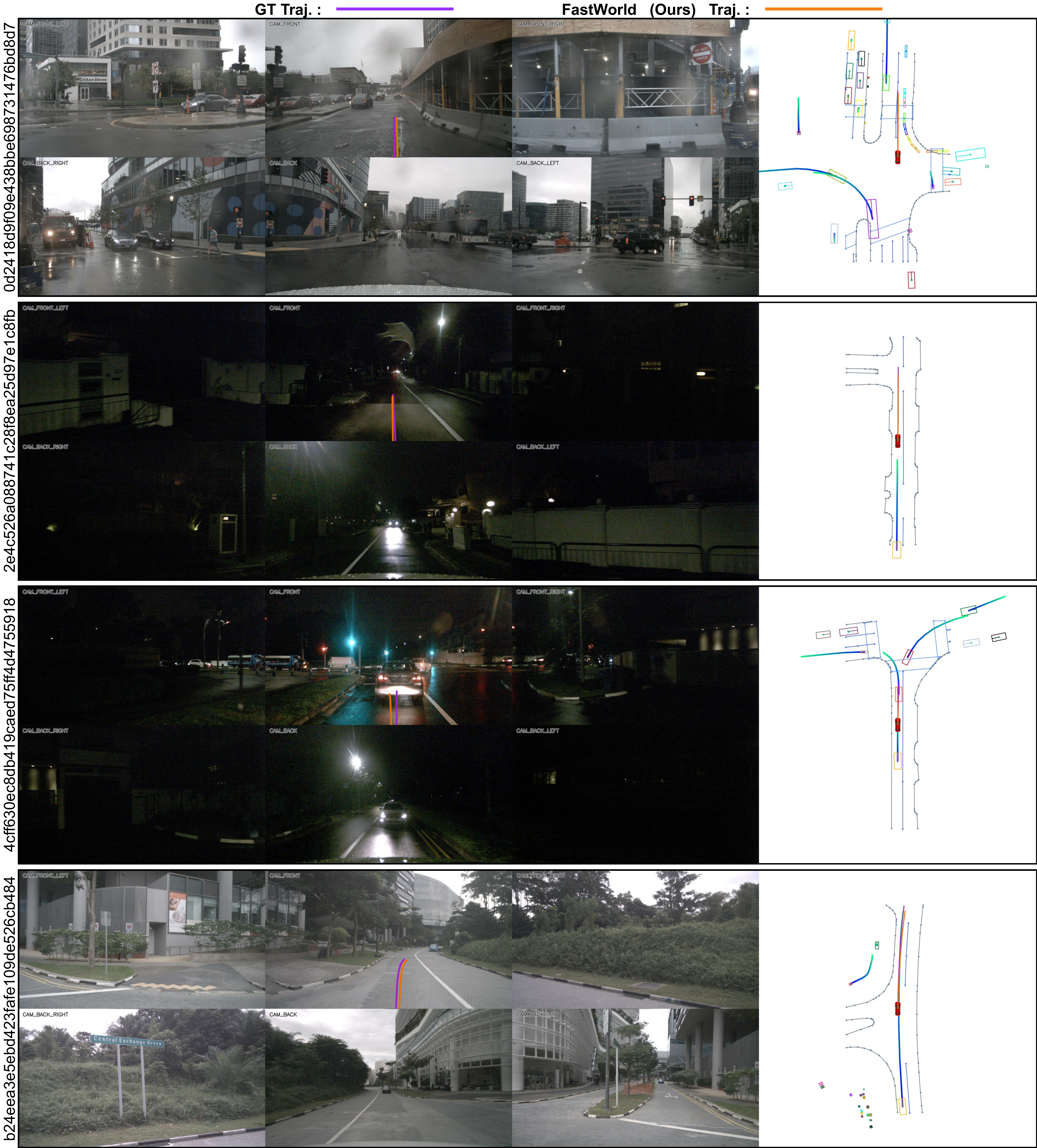}
\caption{\textbf{Visualization examples of \texttt{nuScenes}~\cite{NUSCENES} on straight trajectory.} Left label: sample token. Top label: source of trajectory.}
\label{fig:supp_nus_vis1}
\end{figure*}

\begin{figure*}[!t]
\renewcommand\thefigure{S5}
\centering
\includegraphics[width=1\linewidth]{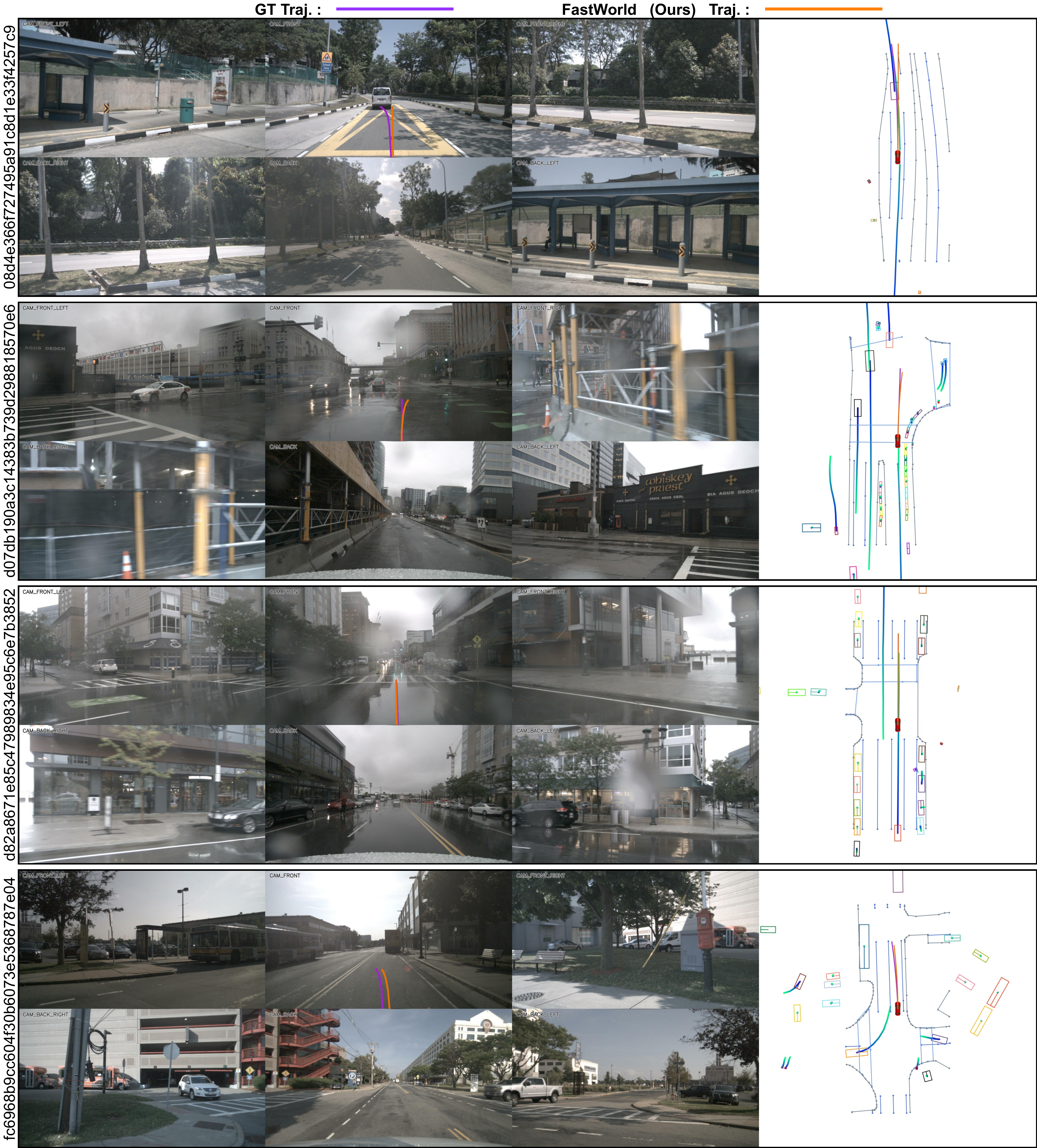}
\caption{\textbf{Visualization examples of \texttt{nuScenes}~\cite{NUSCENES} on straight trajectory.} Left label: sample token. Top label: source of trajectory.}
\label{fig:supp_nus_vis2}
\end{figure*}

\begin{figure*}[!t]
\renewcommand\thefigure{S6}
\centering
\includegraphics[width=1\linewidth]{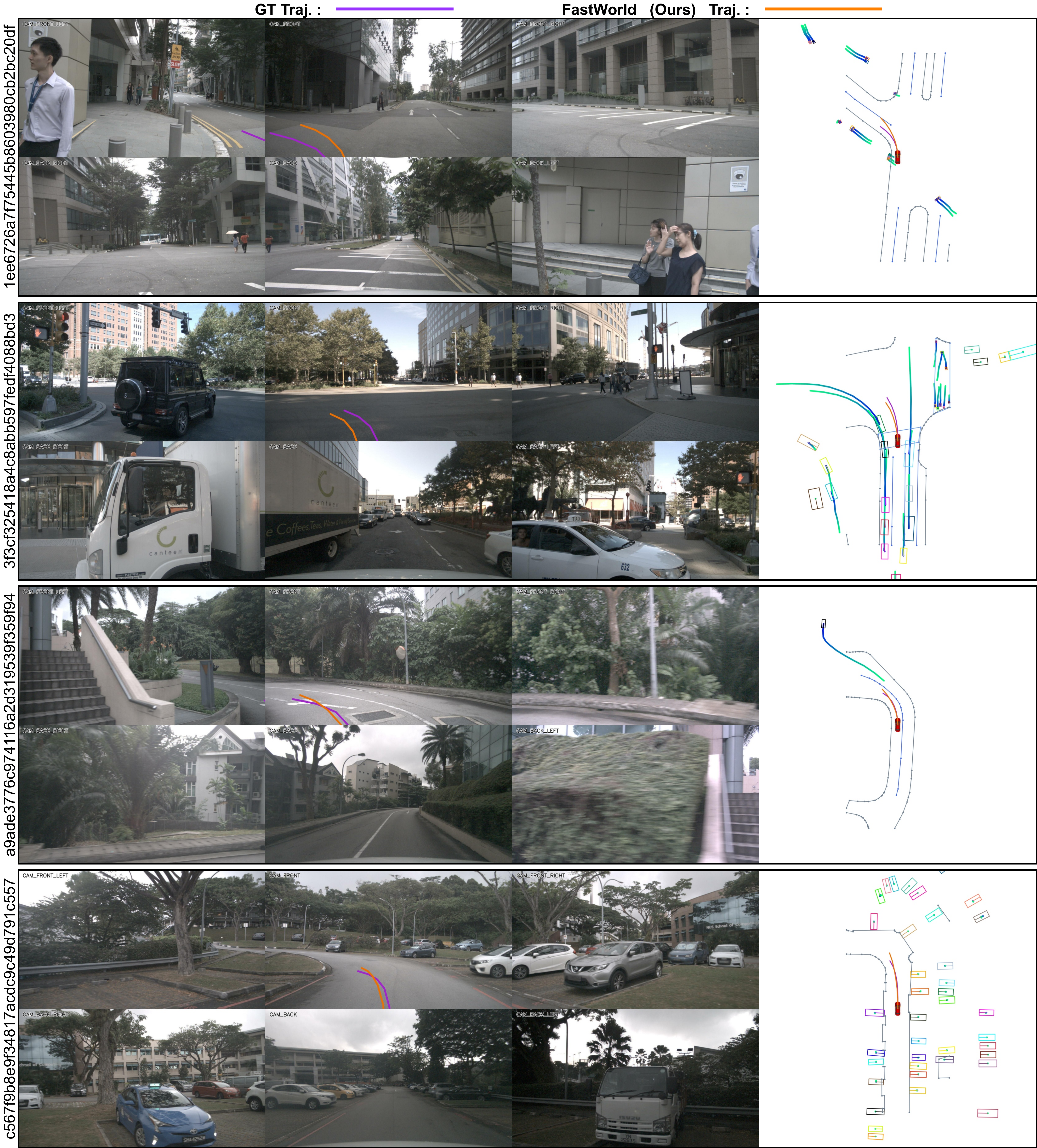}
\caption{\textbf{Visualization examples of \texttt{nuScenes}~\cite{NUSCENES} on left-turn trajectory.} Left label: sample token. Top label: source of trajectory.}
\label{fig:supp_nus_vis3}
\end{figure*}

\begin{figure*}[!t]
\renewcommand\thefigure{S7}
\centering
\includegraphics[width=1\linewidth]{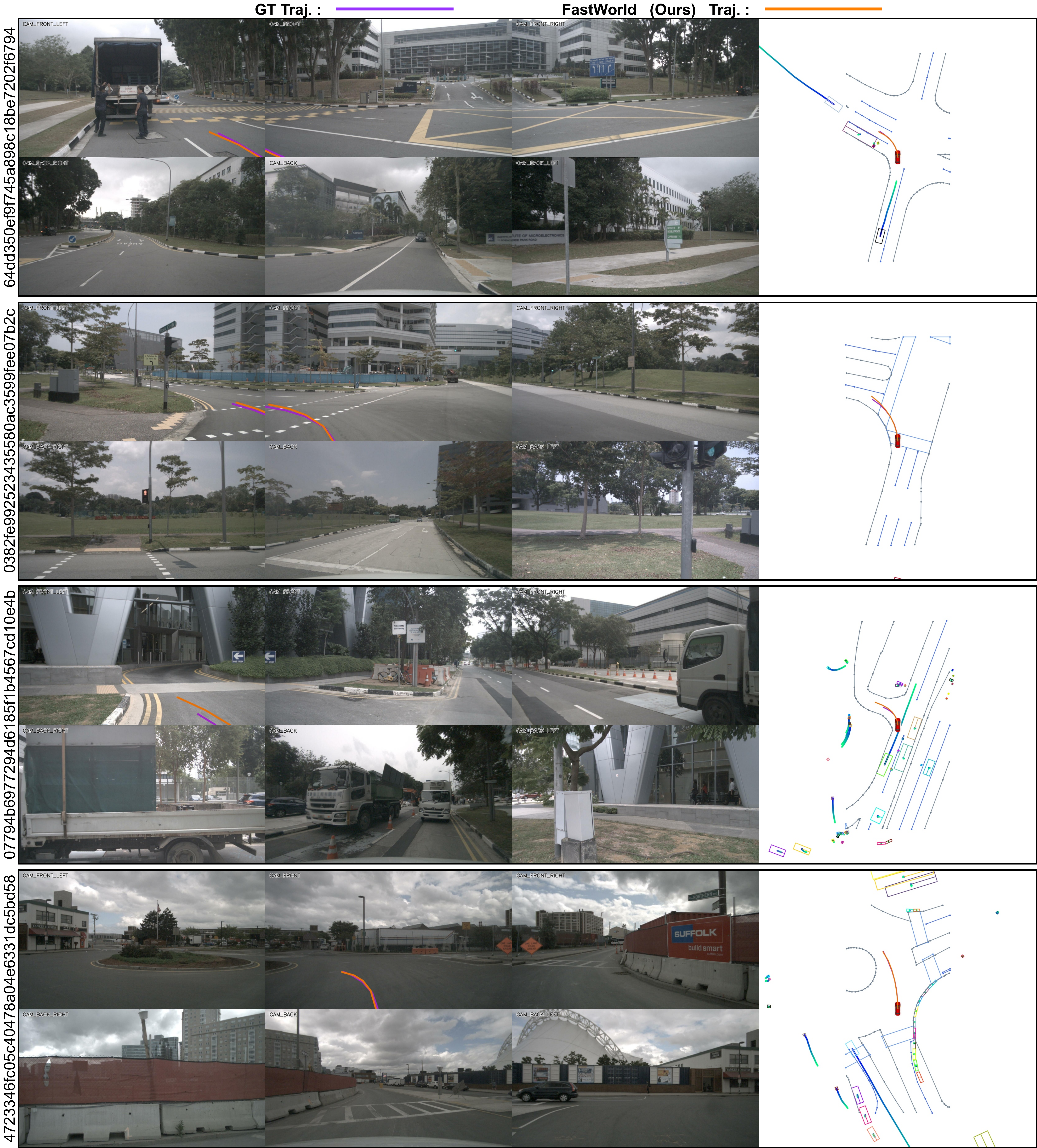}
\caption{\textbf{Visualization examples of \texttt{nuScenes}~\cite{NUSCENES} on left-turn trajectory.} Left label: sample token. Top label: source of trajectory.}
\label{fig:supp_nus_vis4}
\end{figure*}

\begin{figure*}[!t]
\renewcommand\thefigure{S8}
\centering
\includegraphics[width=1\linewidth]{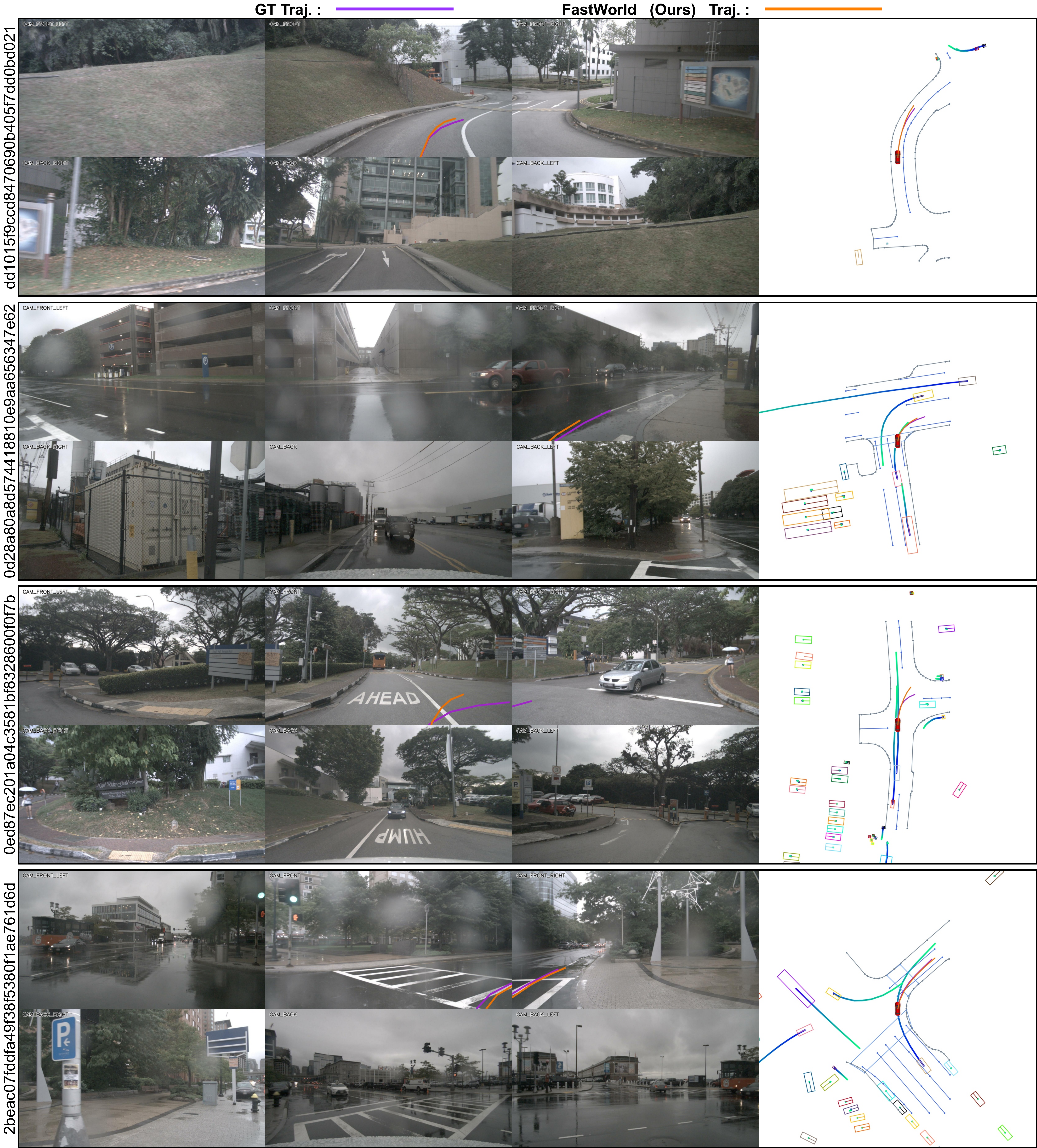}
\caption{\textbf{Visualization examples of \texttt{nuScenes}~\cite{NUSCENES} on right-turn trajectory.} Left label: sample token. Top label: source of trajectory.}
\label{fig:supp_nus_vis5}
\end{figure*}

\begin{figure*}[!t]
\renewcommand\thefigure{S9}
\centering
\includegraphics[width=1\linewidth]{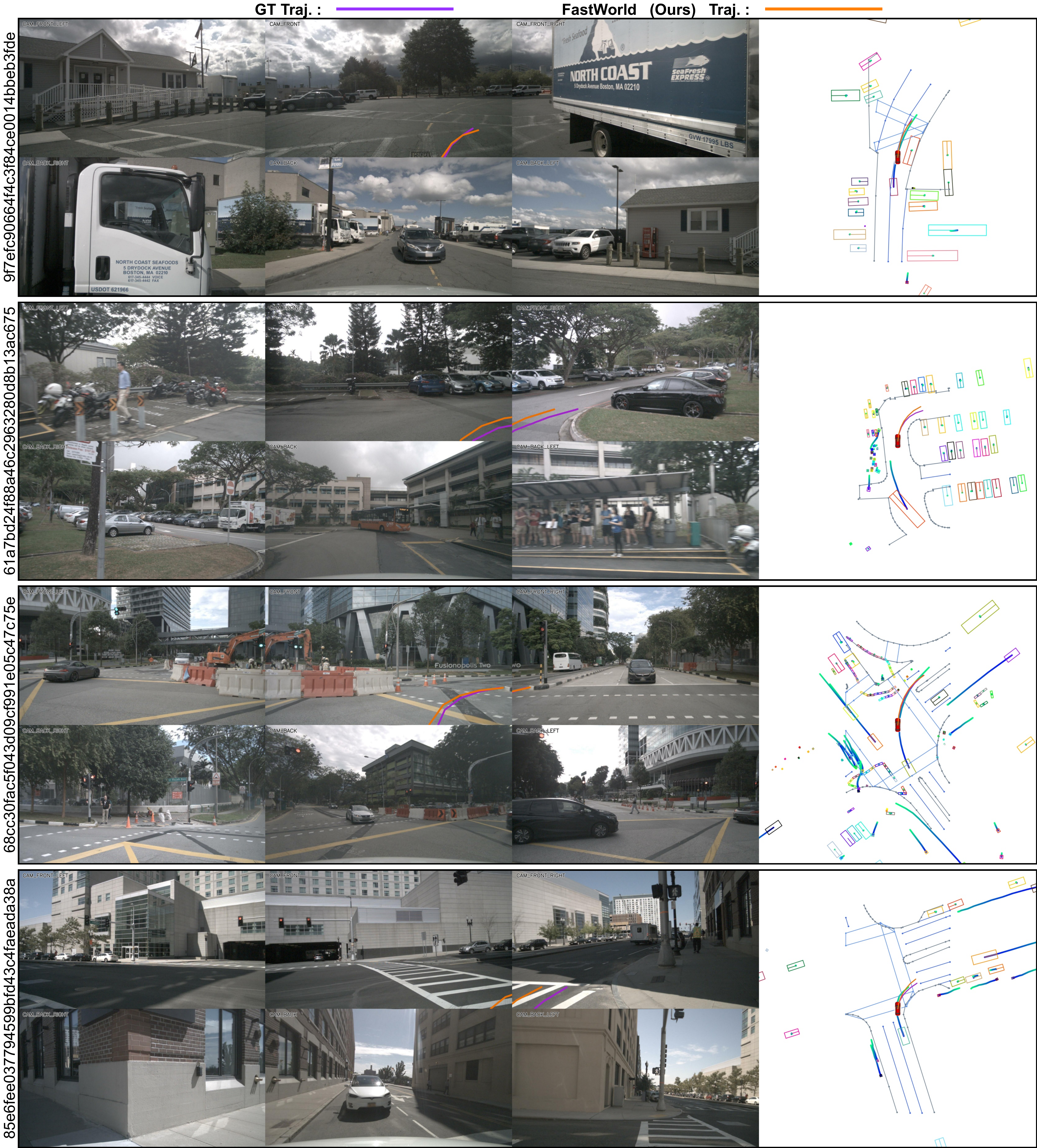}
\caption{\textbf{Visualization examples of \texttt{nuScenes}~\cite{NUSCENES} on right-turn trajectory.} Left label: sample token. Top label: source of trajectory.}
\label{fig:supp_nus_vis6}
\end{figure*}

We have provided more visual examples shown in Fig.~\ref{fig:supp_navsim_vis1}, Fig.~\ref{fig:supp_navsim_vis2}, and Fig.~\ref{fig:supp_navsim_vis3} for NAVSIM~\cite{dauner2024navsim}, as well as Fig.~\ref{fig:supp_nus_vis1}, Fig.~\ref{fig:supp_nus_vis2}, Fig.~\ref{fig:supp_nus_vis3}, Fig.~\ref{fig:supp_nus_vis4}, Fig.~\ref{fig:supp_nus_vis5} and Fig.~\ref{fig:supp_nus_vis6} for nuScenes~\cite{NUSCENES}. Among them, Fig.~\ref{fig:supp_navsim_vis1}, Fig.~\ref{fig:supp_nus_vis1}, and Fig.~\ref{fig:supp_nus_vis2} collect samples where the GT trajectory is straight; Fig.~\ref{fig:supp_navsim_vis2}, Fig.~\ref{fig:supp_nus_vis3}, and Fig.~\ref{fig:supp_nus_vis4} collect samples with left-turn GT trajectories; Fig.~\ref{fig:supp_navsim_vis3}, Fig.~\ref{fig:supp_nus_vis5}, and Fig.~\ref{fig:supp_nus_vis6} collect samples with right-turn GT trajectories.

\noindent \textbf{Visualization of \texttt{NAVSIM}~\cite{dauner2024navsim}.} 
Following the format of Fig.~4 in the main text, we construct visualizations of 4-second trajectories for Drive-JEPA~\cite{wang2026drivejepa}, the GT, and our FastWorld. FastWorld consistently tends to produce conservative trajectories close to GT, while Drive-JEPA~\cite{wang2026drivejepa} yields more aggressive predictions. Representative examples include 2 row 2 col in Fig.~\ref{fig:supp_navsim_vis1}, 3 row 4 col in Fig.~\ref{fig:supp_navsim_vis1}, 1 row 4 col in Fig.~\ref{fig:supp_navsim_vis2}, and 3 row 2 col in Fig.~\ref{fig:supp_navsim_vis3}.

\noindent \textbf{Visualization of \texttt{nuScenes}~\cite{NUSCENES}.} 
We visualize the 3s trajectory predictions of GT and FastWorld, with trajectories projected onto the 6-view camera perspectives. FastWorld generally yields trajectory planning results close to GT.

\clearpage

\end{document}